\documentclass[lettersize,journal]{IEEEtran}
\pdfoutput=1
\usepackage{amsmath,amsfonts}
\usepackage{algorithm}
\usepackage{array}
\usepackage[caption=false,font=normalsize,labelfont=sf,textfont=sf]{subfig}
\usepackage{textcomp}
\usepackage{stfloats}
\usepackage{url}
\usepackage{verbatim}
\usepackage{graphicx}
\usepackage{cite}
\usepackage{times,latexsym}
\usepackage{url}
\usepackage[T1]{fontenc}

\usepackage{multirow}
\usepackage{amssymb}
\usepackage{float}

\usepackage[flushleft]{threeparttable}
\usepackage[utf8]{inputenc}
\usepackage{xcolor}
\usepackage{listings}
\usepackage{enumitem}
\usepackage[english]{babel}
\usepackage{lipsum}
\usepackage{adjustbox}
\usepackage{hyperref}
\usepackage{booktabs}
\usepackage{geometry}
\usepackage[autolanguage]{numprint} 
\usepackage{verbatim}
\usepackage{color}
\usepackage{varwidth}
\usepackage{algpseudocode}
\usepackage{algorithm}
\usepackage{tabularx}
\usepackage{array}

\usepackage{scalerel}
\usepackage{tikz}
\usetikzlibrary{svg.path}

\definecolor{orcidlogocol}{HTML}{A6CE39}
\tikzset{
  orcidlogo/.pic={
    \fill[orcidlogocol] svg{M256,128c0,70.7-57.3,128-128,128C57.3,256,0,198.7,0,128C0,57.3,57.3,0,128,0C198.7,0,256,57.3,256,128z};
    \fill[white] svg{M86.3,186.2H70.9V79.1h15.4v48.4V186.2z}
                 svg{M108.9,79.1h41.6c39.6,0,57,28.3,57,53.6c0,27.5-21.5,53.6-56.8,53.6h-41.8V79.1z M124.3,172.4h24.5c34.9,0,42.9-26.5,42.9-39.7c0-21.5-13.7-39.7-43.7-39.7h-23.7V172.4z}
                 svg{M88.7,56.8c0,5.5-4.5,10.1-10.1,10.1c-5.6,0-10.1-4.6-10.1-10.1c0-5.6,4.5-10.1,10.1-10.1C84.2,46.7,88.7,51.3,88.7,56.8z};
  }
}

\newcommand\orcidicon[1]{\href{https://orcid.org/#1}{\mbox{\scalerel*{
\begin{tikzpicture}[yscale=-1,transform shape]
\pic{orcidlogo};
\end{tikzpicture}
}{|}}}}

\usepackage{hyperref} 

\usepackage{tikz}
\usepackage{textcomp}
\usepackage{hyperref}
\usepackage{lipsum}

\newcommand\copyrighttext{%
  \footnotesize \textcopyright  \textcopyright 2024 IEEE. Personal use of this material is permitted. Permission from IEEE must be obtained for all other uses, in any current or future media, including reprinting/republishing this material for advertising or promotional purposes, creating new collective works, for resale or redistribution to servers or lists, or reuse of any copyrighted component of this work in other works.}
\newcommand\copyrightnotice{%
\begin{tikzpicture}[remember picture,overlay]
\node[anchor=south,yshift=3pt] at (current page.south) {\fbox{\parbox{\dimexpr\textwidth-\fboxsep-\fboxrule\relax}{\copyrighttext}}};
\end{tikzpicture}%
}

\begin{document}

\title{A Comprehensive Evaluation of Neural SPARQL Query Generation from Natural Language Questions}

\author{
\IEEEauthorblockN{
Papa Abdou Karim Karou Diallo \orcidicon{0000-0002-0614-5185}\ \\ Samuel Reyd\orcidicon{0009-0007-7369-5893} and Amal Zouaq \orcidicon{0000-0002-4791-0752} \\ }
\IEEEauthorblockA{ 
\emph{
\{diallokarou28, samuel.reyd, amal.zouaq\}@polymtl.ca 
}} \\
\IEEEauthorblockA{
LAMA-WeST Lab* \thanks{*http://www.labowest.ca/}, Polytechnique Montreal}
}



\maketitle
\copyrightnotice

\maketitle
\begin{abstract}
 In recent years, the field of neural machine translation (NMT) for SPARQL query generation has witnessed significant growth. Incorporating the copy mechanism with traditional encoder-decoder architectures and using pre-trained encoder-decoders and large language models have set new performance benchmarks. This paper presents various experiments that replicate and expand upon recent NMT-based SPARQL generation studies, comparing pre-trained language models (PLMs), non-pre-trained language models (NPLMs), and large language models (LLMs),  highlighting the impact of question annotation and the copy mechanism and testing various fine-tuning methods using LLMs. In particular, we provide a systematic error analysis of the models and test their generalization ability. Our study demonstrates that the copy mechanism yields significant performance enhancements for most PLMs and NPLMs. Annotating the data is pivotal to generating correct URIs, with the "tag-within" strategy emerging as the most effective approach. Additionally, our findings reveal that the primary source of errors stems from incorrect URIs in SPARQL queries that are sometimes replaced with hallucinated URIs when using base models. This does not happen using the copy mechanism, but it sometimes leads to selecting  wrong URIs among candidates. Finally, the performance of the tested LLMs fell short of achieving the desired outcomes.   
\end{abstract}

\begin{IEEEkeywords}
SPARQL query generation; Knowledge base; Copy mechanism; Non Pre-trained and Pre-trained encoders-decoders
\end{IEEEkeywords}

\section{Introduction}
The Semantic Web (SW) provides a framework for storing and organizing structured data, following standards defined by the World Wide Web Consortium (W3C). To access data within a Knowledge Base (KB) on the SW, one must use the SPARQL query language, which can be challenging for non-experts and necessitates a familiarity with the KB's structure and ontology. For greater usability of the SW, it is crucial to develop models that allow users to query KBs easily using natural language. 

One approach to generating queries based on natural language questions is to use neural machine translation (NMT).
This is typically done using an encoder-decoder architecture, where input sentences pass through the encoder to generate a vector that holds their semantics, and the decoder produces tokens at each time step based on the encoder's output and previous tokens. However, classic (without pre-training) NMT architectures have common limitations. First, they require a fixed vocabulary, which cannot be updated without restarting the entire learning process. Therefore, if a new word appears during the testing phase, it is processed as an unknown (\texttt{<unk>}) token. 
Second, it is not trivial for a NMT model to learn to transform natural language into KB elements (URIs), particularly since most elements are seen very few times during training. These limitations are particularly important in the case of NMT-based SPARQL query generation, as unknown words are often elements from the KB, which can lead to irrelevant URIs in the generated queries. To address these limitations, the copy mechanism was proposed   \cite{copy-machanism}, which allows tokens from the input to be directly copied into the output based on a knowledge base vocabulary that includes KB URIs. However, copy-based models require the annotation of KB URIs in the natural language questions. 

The recent development of pretrained language models and their application to SPARQL query generation has opened new potential avenues \cite{lin2022sparql} \cite{reyd2023comprehensive} \cite{tran2021spbert} \cite{huang2021unseen} \cite{huang2021unseen} \cite{naik2023sql}. In fact, Lehmann et al. \cite{lehmann2023language} suggest the use of controlled natural language as a target for Knowledge Graph Question Answering (KGQA) semantic parsing. They hypothesize that pretraining LLMs on textual data can facilitate parsing into controlled natural language for KGQA with limited training data requirements, reducing the cost and effort of collecting high-quality training data. SPARQL can be considered as such a controlled natural language.  
%

Initial experiments on pretrained language models for SPARQL query generation indicate that while they may outperform their non-pretrained counterparts \cite{banerjee2022modern}, they still exhibit some limitations in handling unknown URIs to some degree, but most importantly, they exhibit poor generalization abilities as their performance drops when new question templates are used at test time \cite{reyd2023comprehensive}. With this in mind, while most datasets rely on template-based questions, the ability of neural models trained on template questions to handle natural (formulated by humans, paraphrases) questions remains unexplored. 

In this paper, our goal is to expand upon these experiments and provide a systematic comparison of several models. In particular, we aim to identify the failures of state-of-the-art neural query generators, in terms of SPARQL structures, incorrect URIs, and hallucinated URIs.  We also test models' generalization capabilities with natural - non template-based- questions. Additionally, we include Large Language Models (LLMs) to assess whether they exhibit the same shortcomings. 
In this context, we address the following \textbf{research questions} :
\begin{enumerate}
    \item Does the annotation of KB elements in the natural language questions improve the SPARQL query generation for all models? 
    \item Does the integration of a copy mechanism in NPLMs and PLMs improve the accuracy of the KB elements in the SPARQL queries?
    \item Are Large Language Models (LLMs) effective in this task, and which fine-tuning/prompting technique performs better?
    \item What are the most common generation errors, and what types of tokens are often generated instead of the expected types?
     \item How do models trained on template-based questions perform on naturally reformulated questions?
\end{enumerate}

Our contributions include the following aspects:

\begin{enumerate}
\item{Using annotation, we compare the results of two NPLMs (ConvSeq2Seq and Transformers), two PLMs (BART \cite{BART} and T5 \cite{T5}) and two LLMs (Llama2 \cite{touvron2023llama} and Code Llamav2 7B \cite{roziere2023code}; }
\item{We evaluate the impact of the copy mechanism and question  annotations and experiment with "raw-question" (non-annotated) questions,  "tag-within" questions where we replace natural language elements with their KBs URIs counterparts and  tag-end questions, where we list KB URIs with their labels at the end of the questions;}
\item{
 We experiment with standard fine-tuning and instruction fine-tuning on two Large Language Models (LLMs), namely Llama\cite{touvron2023llama} and Code Llama\cite{roziere2023code} and measure the impact of training data size on the results;} 
\item{We perform a fine-grained analysis of the generation errors with their type distribution for all the models;}
\item{We test the generalization capabilities of the best-performing models using questions reformulated in different settings.}
\end{enumerate}

To our knowledge, such detailed study has not been done yet especially including the performance of large language models, the generalization abilities of all models, and the fine-grained identification of errors made at generation time. 

\section{State of the art}

\paragraph{Non-pre-trained models for SPARQL generation}

Using non-pre-trained encoder-decoders for SPARQL generation has gained a lot of attention in recent years.
\cite{AttentionParsing} presented a method for translating NL statements to SPARQL expressions using an LSTM \cite{LSTM} encoder-decoder model.
\cite{NSPM} proposed an architecture called Neural SPARQL Machines (NSpM), for translating NL expressions to encoded forms of SPARQL queries. The framework involved generating train entries and feeding them to a sequence to sequence model.
\cite{TNTSPA} proposed two new encoder-decoder architectures, ConvSeq2Seq \cite{CNNS2S} and Transformer \cite{Transformer},
on the Monument \cite{NSPM}, LC-QuAD 1.0 \cite{LCQUAD} and DBNQA \cite{DBNQA} datasets. They evaluated their models with BLEU score \cite{BLEU}, query accuracy as well as the F1 score based on the tokens of the candidate and target translation.
\cite{copy-machanism} incorporated a neural copy mechanism on top of ConvSeq2Seq and Transformer architectures and used tagged versions of the questions that identify explicitly entities and relations URIs in the questions. 
The approach was evaluated on the same datasets used by \cite{TNTSPA}, resulting in significant improvements in BLEU scores. They also computed the accuracy of answers obtained after running the generated queries against the knowledge base. 
The approach demonstrated robust performance.

\paragraph{Pre-trained models for SPARQL generation}

Building on the success of encoder-decoder models 
for SPARQL query generation and the recent development of pre-trained models for sequence-to-sequence problems, recent approaches have proposed the use of pre-trained models for SPARQL query generation. 
\cite{pre-trained-special-tokens} proposed an algorithm for identifying entities and relations within a question and subsequently generating query structures with placeholders, which are then populated with URIs in a post-processing step. 
For generating query structures, they used BERT \cite{BERT}, GPT2 \cite{GPT2}, and BART-large \cite{BART}. 
\cite{SGPT} proposed a method for embedding questions using various sources of information, such as POS tagging, and feeding the representation to a GPT2-based model \cite{GPT2}. 
They annotated KB elements in the questions with special tokens depending on the models. These special tokens are replaced by the correct KB elements in a post-processing phase.
\cite{banerjee2022modern} used BART \cite{BART}, T5-small \cite{T5}, T5-base \cite{T5}, and a non-pre-trained model based on Pointer Generation Network \cite{summarization}.  
They added URIs and their labels at the end of the questions (aka "tag-end" annotation). The approach generated several queries using a beam search and ran each query in the order of the beam probabilities, keeping the first one that produced a non-empty answer. The models were evaluated on LC-QuAD 1.0 \cite{LCQUAD} and LC-QuAD 2.0 \cite{LCQUAD2} datasets, using the F1 score on the answers returned by the generated queries. 


\paragraph{Copy mechanism}

Previous approaches have explored the idea of transferring information from the question to the SPARQL query in different ways. Some approaches used placeholders \cite{pre-trained-special-tokens} or cross-attention weights to map words from the question to the query \cite{AttentionParsing}. Recently, explicit copy mechanisms inspired by CopyNet \cite{CopyNet} and Pointer-Generator Networks (PGN) \cite{summarization} have been incorporated into encoder-decoder architectures. These mechanisms allow the decoder to generate tokens based on probabilities derived from the decoder's logits or by copying tokens from the input. CopyNet \cite{CopyNet} uses a learned weight matrix to calculate the probability of copying each word from the input. PGN \cite{summarization} computes a copy probability and copy scores based on cross-attention weights, combining them with generation scores using the copy probability. \cite{banerjee2022modern} directly used the PGN \cite{summarization} architecture, whereas \cite{copy-machanism} proposed a modified version where the URIs are masked in the input sequence.  

In this work, we plan to unify the evaluation of both NPLM and PLMs under the same question annotation and the same copy / no copy settings. To our knowledge, none of the available approaches have incorporated a copy mechanism in architectures such as BART \cite{BART} or T5 \cite{T5}.

\paragraph{Large Pretrained Language Models}
Finally, recent works have also explored Large Language Models (LLMs) for SPARQL query generation \cite{muennighoff2022sgpt}  \cite{yang2023llm} \cite{kovriguina2023sparqlgen}. For instance, \cite{muennighoff2022sgpt} propose SGPT, an extension of SBERT that enhances GPT models for semantic search. 
 \cite{luo2023chatkbqa} introduce ChatKBQA, a novel KBQA framework built on fine-tuned open-source LLMs, which combines generation and retrieval for improved performance on standard KBQA datasets. In particular, the approach generates logical forms before retrieving entities and relations, followed by a conversion to SPARQL queries. 
Kovriguina et al. \cite{kovriguina2023sparqlgen} introduce SPARQLGEN, a generative approach for generating SPARQL queries using GTP-3 \cite{GPT3}, leveraging various types of contexts in the prompt to influence query generation performance. 
Their approach is similar to ours for LPLMs except that we use only a single prompt structure and use some of the latest LPLMs Llama2 \cite{touvron2023llama} and Code Llamav2 7B \cite{roziere2023code}.  
While all these approaches leverage large pre-trained language models, they do not focus on their performance in out-of-distribution settings and do not analyze their limitations in detail.  

\section{Methodology}

\subsection{Task and Data Format} \label{section:data}


Each dataset is a standard, publicly available dataset, composed of a set of question-query pairs called entries. The question is formulated in English and the target is a SPARQL query. All the datasets are generated automatically using templates. These templates match question-query structures with placeholders that are later filled with specific URIs. We call this matching pair a global template. Since some question structures can be associated with several query structures and vice versa, we also refer to question templates and query templates as the isolated question or query structures. For example, the global template Question: "what is the \texttt{<1>} of \texttt{<2>} ?" / Query: \texttt{select distinct ?uri where \{<2> <1> ?uri\}} generated the following Entry Question: "what is the office of richard coke ?" and its associated Entry Query: \texttt{select distinct ?uri where \{ dbr:Richard\_Coke dbp:office ?uri \}}.

\subsubsection{Annotations} \label{section:annotation}
To identify the impact of annotations on models' performance, we experiment with three schemes. \paragraph{"Raw questions"} Questions without annotation are designated as "raw-question" in our results.
\paragraph{"tag-within" questions}  
"Tag-within" questions are generated by substituting the natural language words within the placeholders of the question template with the corresponding URIs and literals extracted from the placeholders of the query template. For instance, the "tag-within" question associated to the example "which person has opponent ike clanton ?" is "who is the \texttt{<<dbo:Person>>} whose \texttt{<<dbo:opponent>>} is \texttt{<<dbr:Ike\_Clanton>>} ? ". 
In contrast to the approach proposed by \cite{banerjee2022modern} and \cite{copy-machanism}, this study includes not only URIs but also literals. For example, the question "Give me organization that contains the word zollkriminalamt in their name" expects the query to filter the organization names based on the appearance of the string "zollkriminalamt".
This modification is motivated by the fact that literals are also KB elements that should be transferred directly from the question to the query, similarly to URIs.

\paragraph{"tag-end" questions} 
In tag-end questions,  the list of KB URIs available in the SPARQL query is appended to the question. Instead of replacing the corresponding natural language tokens, URIs  are randomly placed at the end of the question, together with their corresponding English label. The inclusion of labels next to the URIs improves the understanding of the semantics of the URIs.
Each pair of URI/label is separated by a \texttt{<sep>} token. The "tag-end" question associated to the previous example is "who is the person whose opponent is ike clanton ? \texttt{<sep> <<dbr:Ike\_Clanton>> ike clanton <sep> <<dbo:opponent>> opponent <sep> <<dbo:Person>> person}".

\subsection{Base model architectures}

\paragraph{Non-pre-trained models}
We first tested non-pre-trained encoder-decoder models. Following \cite{TNTSPA} and \cite{copy-machanism}, we used a Transformer model \cite{Transformer} and a ConvSeq2Seq model \cite{CNNS2S}. We re-ran the experiments presented in these two papers ("raw-question" and "tag-within" questions on LC-QuAD 1.0 \cite{LCQUAD} and DBNQA \cite{DBNQA}) and extended the experiments with an additional dataset (LC-QuAD 2.0 \cite{LCQUAD2}) and an additional data annotation method ("tag-end" questions).
These results shall demonstrate the limitations explained in the introduction and set a baseline for comparison with the following models.

\paragraph{Pre-trained models}
To compare our results with \cite{banerjee2022modern}, we used BART-base \cite{BART} and T5-small \cite{T5} as pre-trained encoder-decoder architectures. We also compare these architectures with the non-pre-trained ones in the same experimental settings to show the impact of pre-training. For SPARQL schema elements and URIs, we followed the encoding described by \cite{banerjee2022modern}. Each SPARQL schema element and each URI prefix is considered as a  special token. With T5, we used the sentinel tokens and with BART, we added new tokens to represent these elements. 

\paragraph{Large Pre-trained models}
In order to evaluate Large Language Models,  we made use of Ludwig\footnote{https://ludwig.ai/latest/}, an open-source framework  for training and deploying machine learning models. Ludwig streamlines the intricate process of constructing, training, and deploying these models. We experimented with two LLMs:  Llama \cite{touvron2023llama}, as it is among the top-performing models in several NLP tasks, and Code Llama\cite{roziere2023code}) which is specialized for code generation. Using Ludwig, we fine-tuned these models in two ways: first, we performed fine-tuning with just the input (NL question) and output (SPARQL query). Then, we used a prompt with an instruction explaining the task, followed in the input sequence by the NL question and the SPARQL query as output. We also provided the KB elements needed to generate the SPARQL query in the latter method. Providing the URIs through the prompt's instruction allows us to simulate the tag-end setting as an annotation method used with the PLMs and the NPLMs. 

\subsection{Copy mechanism}

As stated previously, the main limitation of neural query generators is the accurate generation of URIs in the SPARQL queries, and this is even more important when these URIs are unknown.  
Thus our hypothesis is that the SPARQL query generation task can be decomposed into a query structure generation, followed by a copy of the URIs at some specific positions. 
We extend the use of the copy mechanism  \cite{copy-machanism}, which is based on only copying elements from a KB vocabulary and masking these elements to the encoder and decoder blocks. Our extension includes literals in the KB vocabulary and the addition of the copy to the pre-trained models' architectures. This is further explained in the following sections.


\subsubsection{Vocabularies}
We define three distinct vocabularies: 

\begin{itemize}
    \item The English natural language vocabulary, denoted as $W$, composed of the questions' tokens .
    \item The SPARQL vocabulary, denoted as $S$, which includes SPARQL keywords.
    \item The KB vocabulary, denoted as $K$, which includes URIs (classes, properties, and resources) as well as literals. 
\end{itemize}

Each entry of our datasets is composed of a question with tokens from $W$ and a query with tokens from $K \cup S$. Our annotated versions of the dataset 
tag the question elements with URIs from $K$ that appear in the query. In the "tag-end" setting, the size of $W$ might be slightly larger since the labels used next to the URIs are not necessarily the same as those used in the original question. That is because we use the URI's English label defined in the KB. 
On the opposite, in the "tag-within" version, the size of $W$ is massively reduced since the words used to describe specific URIs in the question are replaced by tokens from $K$.

\subsubsection{Description}
The copy mechanism computes, at each generation step, a probability of copy for the next token. This probability weights the generation score and copy score of the next token. For each word of the target vocabulary, the generation score is based on the logits of the decoder. The copy score is based on the cross-attention weight between the next predicted token and each token of the input. We arbitrarily chose to use the last attention head for every model. Unlike PGN \cite{summarization}, this implementation of the copy mechanism is designed to only copy tokens from a specific vocabulary, here the KB vocabulary $K$.
When the input sentence is provided to the encoder, the words from this vocabulary are masked. Then, the copy scores are only computed for these masked tokens. 

More formally, at generation step $i$, the generation probability of a token in $S \cup K$ is computed as follows:

\begin{equation*}
    p_{t_i} = p_{copy} \times p_C + (1 - p_{copy}) \times p_G
\end{equation*}

Where $p_{t_i}=P(t_i|w_{0:m};t_{0:i-1})$, with $t_i$ being the i-th token generated, $t_{0:i-1}$ being the tokens previously generated, and $w_{0:m}$ being the original input tokens before masking. 

$p_{copy}$  is the probability of using copy instead of generation, $p_C$ the probability of copying a token, $p_G$ is the probability of generating a token according to the decoder.

We can express $p_G$, $p_C$ and $p_{copy}$ as such:

\begin{flalign*}
    & p_G = \begin{cases}
                \sigma_S(\text{DEC}_{i}(t_i|\tilde{w}_{0:m};t_{0:i-1})) & \forall t_i \in S \\
                0 & \forall t_i \notin S
            \end{cases} \\
    & p_C = \begin{cases}
                \sigma_{K\cap w_{0:m}}(A_{k,i}) &  \forall t_i \in K \; \text{s.t.}\ \exists k : w_k=t_i\\
                0 & \text{otherwise}
            \end{cases} \\
    & p_{copy} = \sigma( \text{DEC}_{i}(t_i|\tilde{w}_{0:m};t_{0:i-1})) \times B)
\end{flalign*}

With $\tilde{w}_{0:m}$ being the tokens given as input to the encoder where the words from $K$ have been masked, $\text{DEC}_{i}$ being the logits of the decoder at timestep $i$, $A_{k , i}$ is the cross-attention weight computed by the encoder-decoder between the k-th term of the input and the i-th term of the predicted output, $B \in \mathbb{R}^{|S| \times 1}$ is a matrix with weights learned during training, $\sigma(\cdot)$ is the sigmoid function and $\sigma_X(\cdot)$ is the softmax function computed over set $X$.

In the end, with this copy mechanism, the encoder vocabulary is $W$, the decoder vocabulary is $S$, and only tokens from $K$ can be copied from the question to the query.

\section{Experiments}\label{section:experiment}


In our experiments, we combined various parameters, including six different base model architectures, the addition or removal of the copy mechanism for pre-trained and non-pre-trained models, two types of question annotations, and data from three different datasets. We conducted eight experiments with Large Language Models (LLMs) using two different models and two fine-tuning strategies (standard fine-tuning and instruction fine-tuning) with varying sample proportions. For Pre-trained Language Models (PLMs) and Non-Pre-trained Language Models (NPLMs), we conducted 56 experiments. We carried out 56 tests, comprising ten adaptations of existing studies and 46 original results, which we will discuss in more detail in the following sections.

\subsection{Datasets} 
We experiment with three datasets. Datasets statistics can be seen in Table \ref{tab:sizes} and  Table \ref{tab:voc}.  The vocabulary sizes for the three datasets are presented in Table \ref{tab:voc}. We also report the size of the out-of-vocabulary (OOV) tokens as it highlights the difficulties mentioned in the introduction. We can notice that the set $S$ does not include OOV tokens whereas the sets $W$ and $K$ often feature a significant amount of OOV tokens.

\paragraph{LC-QuAD 1.0} The LC-QuAD 1.0 (referred to as LCQ1 for short in tables) dataset \cite{LCQUAD} 
includes two versions of the question. The first one is the automatically generated question based on templates (aka generation templates). The second is a human reformulation. For example, the question "\texttt{how many movies are there whose director is Stanley Kubrick?}" is reformulated as "\texttt{how many movies did Stanley Kubrick direct ?}". The dataset also includes, for each question, a template id  corresponding to the query template that generated the SPARQL query. To apply our tagging methodology, we identified the question templates for each question 
and extracted the global templates associated to it. We found 35 global templates, 23 question templates and 34 query templates (see section III.A for a definition). 
\paragraph{LC-QuAD 2.0} The LC-QuAD 2.0 (referred to as LCQ2 for short in tables) dataset \cite{LCQUAD2} is  richer than LC-QuAD 1.0 
since it also includes literals and filtering operations (i.e. the \texttt{FILTER} option in the SPARQL query). 
It also includes reformulated questions. We can see in Table \ref{tab:voc} that the size of the SPARQL vocabulary $S$ is much bigger than in LC-QuAD 1.0 \cite{LCQUAD} and even bigger than in DBNQA \cite{DBNQA} despite the overall size difference. Another difficulty is that LC-QuAD 2.0 \cite{LCQUAD2}, by adding filtering operations, is the only one of our datasets to feature literals. Hence, it is the most challenging of our datasets. We found 34 global templates, 27 question templates and 33 query templates. 

\paragraph{DBNQA} We extracted a subset of The DBNQA dataset \cite{DBNQA}, which is far more massive than our two other datasets. The reason of using only a subset is that the original DBNQA dataset does not include any information about any of the generation templates associated with each entry. To be able to tag the questions, we used  a subset of this dataset for which we could identify the global templates. 
These global templates were then used to generate the "tag-within" questions. We found 512 global templates, 507 question templates and 157 query templates. 


\begin{table}[]
\caption{Size of datasets.}
    \centering
    \small
    \begin{tabular}{cccc}
        \toprule
         & \textbf{LCQ1} & \textbf{LCQ2} & \textbf{DBNQA} \\
         \midrule
        Total & 5000 & 30225 & 382794 \\
        Train & 4000 & 21761 & 306236 \\
        Val & 500 & 2418 & 38279 \\
        Test & 500 & 6046 & 38279 \\
        \bottomrule
    \end{tabular}
    \label{tab:sizes}
\end{table}

\begin{table}
    \caption{Size of vocabularies.}
    \centering
    \begin{tabular}{cccccc}
        \toprule
        & \textbf{Total} & \textbf{Train} & \textbf{Val} & \textbf{Test} & \textbf{OOV} \\
        \midrule
        \multicolumn{6}{c}{NL vocabulary $W$} \\
        \hline
        LCQ 1 & 6141 & 5481 & 1631 & 1644 & 368 \\ 
        LCQ 2 & 25975 & 21863 & 5602 & 10124 & 5751 \\ 
        DBNQA & 95974 & 89128 & 32267 & 32440 & 3836 \\\hline
        \multicolumn{6}{c}{SPARQL vocabulary $S$} \\
        \hline
        LCQ 1 & 13 & 13 & 13 & 13 & 0 \\ 
        LCQ 2 & 45 & 45 & 45 & 45 & 0 \\ 
        DBNQA & 33 & 33 & 33 & 33 & 0 \\ 
        \hline
        \multicolumn{6}{c}{KB vocabulary $K$} \\
        \hline
        LCQ 1 & 4751 & 4150 & 1068 & 1065 & 318 \\ 
        LCQ 2 2.0 & 34801 & 27949 & 5413 & 10993 & 7523 \\ 
        DBNQA & 157672 & 142985 & 36845 & 37130 & 7998 \\
        \bottomrule
    \end{tabular}
    \label{tab:voc}
\end{table}

\paragraph{Dataset answers} 
The datasets used in this study do not provide the answers to the question-query pairs they contain. To address this issue, we enriched these datasets with the expected answers for each query in the three datasets. 
To extract the answers, 
we used a DBpedia SPARQL endpoint for LC-QuAD 1.0 and DBNQA. The DBpedia endpoint was based on a 2016 dump, which was the version available at the time of the LC-QuAD 1.0 publication. We used the same version for DBNQA due to the unavailability of dumps for the intended 2018 version. We also implemented a local Wikidata 2021 endpoint for LC-QuAD 2.0 to reduce the number of queries returning empty answers on other endpoint versions and for speed considerations at test time.
All reported results are computed only on the subsets of non-empty answers of the test sets. This is especially important since this can lead to inflated performances when models provide incorrect queries that return empty answers, yet still receive positive scores with answer-based metrics if the gold standard also includes empty answers.   After discarding empty answers, we keep 100\% of the LC-QuAD 1.0 test set, 96.20\% of the LC-QuAD 2.0 test set and 80.41\% of the DBNQA test set.

\subsection{Models hyperparameters}

For our experiments, we chose not to focus on the optimization of the hyperparameters but on the comparison between approaches. Therefore, we based our experiments on the experimental settings of state-of-the-art approaches and based on computational resources limitations.

For the Transformer and the ConvSeq2Seq architectures, we followed \cite{TNTSPA} hyperparameters. 
For the BART and T5 models, we followed \cite{banerjee2022modern} recommendations. 
We also followed the recommendations from \cite{copy-machanism} and \cite{banerjee2022modern} to train the models. We chose to only experiment with T5-small since \cite{banerjee2022modern} reported better performance with this version on LC-QuAD 1.0 and similar performance between T5-small and T5-base on LC-QuAD 2.0. We also chose BART-base to compare our results with those of \cite{banerjee2022modern}.  Details of all the hyper parameters are shown in Table \ref{tab:hyperparameters}.

We used batch sizes of 32, 16 and 5 for the non-pre-trained models (respectively for LC-QuAD 1.0, LC-QuAD 2.0 and DBNQA). We only adjusted the batch size of the Transformer for DBNQA to 32 because it helped raise performance. For pre-trained models, we used batch sizes of 16, 8 and 5 respectively for LC-QuAD 1.0, LC-QuAD 2.0 and DBNQA. 
For non-pre-trained experiments, we trained our models for 500, 150 and 50 epochs (respectively for LC-QuAD 1.0, LC-QuAD 2.0
and DBNQA). For pre-trained models, 
we used 200, 50 and 20 epochs. Each reported result is the mean of three different runs with the different random seeds. We train the model train for the number of specified epochs for each run and keep the model with the best validation loss for testing.  The training is done with teacher forcing, i.e. the decoder is supposed to predict the next token of the gold standard query. 
For generation at test time, we use greedy decoding.

\begin{table}
\caption{Hyper parameters for our models.}
\centering
\begin{threeparttable}
\begin{tabular}{ccccccc}
    \toprule
    \textbf{Model} & \textbf{Layers}  & \textbf{H.Units} &  \textbf{Optim.} &  \textbf{LR} & \textbf{LRS.} &  \textbf{Dropout} \\
    \midrule
     CNN & 15 & 512 & SGD & 0.5 &  No & 0.2 \\
     Transf. & 6 & 1024 & Adam & 0.0005 & No & 0.3 \\
     BART & 6 & 768 & Adam & 0.000015 & PDSW & 0.1 \\
     T5 & 6 & 512 & Adam & 0.0015 & PDSW & 0.1 \\
    \bottomrule
\end{tabular}
    \begin{tablenotes}
     \item[]  PDSW: Polynomial decay schedule with warmup. Transf: Transformer. \\
     CNN: ConvSeq2Seq.  LRS: Learning Rate Scheduler 
     \end{tablenotes}
\end{threeparttable}
\label{tab:hyperparameters}
\end{table}



\subsection{Metrics}

To evaluate the performance of our models, we use three metrics. 
First, we report the BLEU-score \cite{BLEU}, which is a popular NMT metric that compares the predicted query to the gold standard query at the token level. We also report two question-answering metrics. The first one is the accuracy of the produced answer. Each predicted query is ran against the KB and we compare the returned answer to the answer of the gold standard. This is the most relevant metric to compare to studies such as \cite{banerjee2022modern}, even though we did not use the strategy of keeping the first non-empty answer amongst the top-n generated queries. Indeed, our models only generate one query. Finally, we compute the F1-score of the answers. 
We then average the F1-scores of each entry for each of the three runs to get the F1-score of the model on a given test set. 

\section{Results}

We report results rounded to integer to facilitate comparison between results within the same tables. The results for non-pre-trained models can be found in Table \ref{tab:scratch_original_results}. 

\subsection{Non-pre-trained models}

\begin{table*}
\caption{Results for non-pre-trained models.}
\centering
\begin{threeparttable}
\begin{tabular}{c|c|ccc|ccc|ccc|ccc}
\multicolumn{2}{c}{} & \multicolumn{6}{c}{Transformer} \vline & \multicolumn{6}{c}{ConvSeq2Seq} \\
\cline{3-14}
\multicolumn{2}{c}{} &  \multicolumn{3}{c}{Base} \vline & \multicolumn{3}{c}{Copy} \vline &  \multicolumn{3}{c}{Base} \vline & \multicolumn{3}{c}{Copy}\\
\cline{3-14}
\multicolumn{2}{c}{} & BLEU & Acc. & F1 & BLEU & Acc. & F1 & BLEU & Acc. & F1 & BLEU & Acc. & F1 \\
\hline
\multirow{3}{*}{LCQ 1} 
& raw-question   & 69 & 33 & 37 &    /  &    /  &   /      
        & 69 & 23 & 27 &    /  &    /  &   /  \\  
        
& tag-within & 82 & 41 & 44 & 99 & 95 & 95    
             & 78 & 28 & 32 & 99 & 96 & 96 \\ 
             
& tag-end   & 74 & 38 & 41 & 49 & 1.8  & 1.8     
           & 75 & 25 & 28 & 90 & 77 & 78 \\ 
\hline
\multirow{3}{*}{LCQ 2} 
& raw-question   & 57  & 0.8 & 1.1 &  /  &  /  &  /      
        & 76  & 9.8 & 11  &  /  &  /  &  /  \\  
        
& tag-within  & 59  & 1.5 & 1.6 & 82  & 69  & 69    
              & 76  & 8.8 & 10  & 89  & 69  & 69 \\ 
              
& tag-end  & 66  & 1.4 & 1.7 & 57 & 2.2 & 2.2    
           & 78  & 14  & 15  & 90 & 66  & 66  \\ 
\hline
\multirow{3}{*}{DBNQA}   
& raw-question  & 58 & 47 & 47 &  / & / & /      
       & 71 & 52  & 52  &  / & / & /   \\ 
       
& tag-within  & 58 & 45 & 45 & 93 & 86 & 87    
              & 75 & 59  & 60  & 97 & 92 & 92 \\ 
              
& tag-end   & 60 & 1.7 & 2.2 & 60 & 7.0 & 7.2   
            & 92 & 76  & 76  & 96 & 88  & 88 \\ 
\end{tabular}
\label{tab:scratch_original_results}
\begin{tablenotes}
     \item[] All values in the table are percentages.
  \end{tablenotes}
\end{threeparttable}
\end{table*}

\paragraph{Reproduction results} \label{sec:reproduction-scratch}

Parts of our results reproduce existing studies. For models without and with copy, results have already been reported on LC-QuAD 1.0 and DBNQA  without annotation \cite{TNTSPA,copy-machanism} and with "tag-within" questions \cite{copy-machanism}.  
We can observe that our results slightly improve those reported by \cite{TNTSPA, copy-machanism} for LC-QuAD 1.0 with Transformer and ConvSeq2Seq models with and without the copy mechanism. We also report similar results for DBNQA with ConvSeq2Seq with and without the copy mechanism. 


\paragraph{Annotation impact}
For almost all of our results, we can see that the question annotation improves the performance.
Performances diminish a little only for Transformer on DBNQA and for ConvSeq2Seq on LC-QuAD 2.0 with "tag-within" questions compared to no annotation ("raw-question"). Except for these two counter examples, we see a consistent improvement due to annotation. 

In the copy-based architectures, we can note that Transformers reach good performances with "tag-within" questions and completely fail with "tag-end" questions. ConvSe2Seq models reach good performances with both settings. When using the copy mechanism, we don't run experiments on "raw-question" questions because they don't include URIs to copy.

\paragraph{Impact of the copy mechanism}
The copy mechanism almost always has a huge positive impact on performance. The impact of copy is always more significant with "tag-within" questions than with "tag-end" questions. Finally, we can notice that the DBNQA dataset is the one that benefits the less from the copy mechanism. This is probably because the huge amount of data allows the non-pre-trained models to learn already well enough without the copy mechanism. There is, however, still a significant jump in performance compared to non-copy models. 

%
%

\subsection{Pre-trained models}
\paragraph{Base Models}
We compare BART and T5 and reproduce the results of \cite{banerjee2022modern} on top of additional experiments. We can note that T5 consistently outperforms BART, and that "tag-within" questions always imply similar or better performance than "tag-end" questions as shown in Table \ref{tab:pretrained_original_results}. We can also see that for the "tag-end" questions without the copy mechanism, our results are better or close to the results reported by \cite{banerjee2022modern} on LC-QuAD 1.0 and LC-QuAD 2.0. Our results for BART are much better than what they report, whereas we obtain a drop of 3 points for T5 on LC-QuAD 2.0. This might be due to the fact that we include the literals at the end of the questions, contrary to \cite{banerjee2022modern}. Finally, our generation strategy is different. We evaluate the greedy generation of our models whereas \cite{banerjee2022modern} kept the top-10 beam-generated queries, ran each of them on the endpoint, and only evaluated the first one to return a non-empty answer.

\begin{table*}
\caption{Results for pre-trained models.}
\centering
\begin{threeparttable}
\begin{tabular}{c|c|ccc|ccc|ccc|ccc}
\multicolumn{2}{c}{} & \multicolumn{6}{c}{BART} \vline & \multicolumn{6}{c}{T5} \\
\cline{3-14}
\multicolumn{2}{c}{} &  \multicolumn{3}{c}{Base} \vline & \multicolumn{3}{c}{Copy} \vline &  \multicolumn{3}{c}{Base} \vline & \multicolumn{3}{c}{Copy}\\
\cline{3-14}
\multicolumn{2}{c}{} & BLEU & Acc. & F1 & BLEU & Acc. & F1 & BLEU & Acc. & F1 & BLEU & Acc. & F1 \\
\hline
\multirow{3}{*}{LCQ 1}      
& raw-question   & 84 & 72 & 75 & / & / & /     
        & 81 & 54 & 57 & / & / & /  \\ 
& tag-within  & 96 & 87 & 87 & 99 & 96 & 96     
              & 99 & 97 & 97 & 99 & 96 & 96  \\ 
& tag-end  & 94 & 84 & 85 & 97 & 94 & 93     
           & 98 & 96 & 96 & 98 & 92 & 93  \\ 
\hline
\multirow{3}{*}{LCQ 2}         
&  raw-question   & 72 & 4.7 & 4.8 &    / & / & /      
         & 79 & 13  & 14  &    / & / & /   \\ 
         
& tag-within  & 87 & 84 & 84 & 95 & 85 & 85     
              & 95 & 94 & 94 & 90 & 72 & 72  \\ 
              
&tag-end  & 88 & 84 & 84 & 89 & 72 & 72     
          & 90 & 90 & 90 & 87 & 58 & 58  \\ 
\hline
\multirow{3}{*}{DBNQA}          &   raw-question        & 87 & 61 & 61 &    /  &    /  &   /      
& 88 & 60 & 60 &    /  &    /  &   /  \\  
& tag-within        & 94 & 79 & 79 & 94 & 90 & 90    
& 97 & 97 & 97 & 76 & 54 & 56\\  
& tag-end       & 94 & 82 & 83 & 94 & 91 & 91    
& 97 & 94 & 94 & 24 & 16 & 17\\  
\end{tabular}
\label{tab:pretrained_original_results}
\begin{tablenotes}
     \item[] All values in the table are in percentages
  \end{tablenotes}
\end{threeparttable}
\end{table*}

\paragraph{Copy-based Models}
We can note that BART benefits from the copy mechanism much more than T5. Except for LC-QuAD 2.0 with "tag-within" questions, the copy mechanism always allows a rise in performance for BART. The case of LC-QuAD 2.0 with "tag-within" questions coupled with copy-based models shows a specific difficulty that is discussed in Section \ref{sec:discussion}.
On the contrary, the copy mechanism only allows a slight rise in performance for T5 with the LC-QuAD 1.0 dataset. However, it considerably lowers the results with the "tag-end" questions on both the LC-QuAD 2.0 and DBNQA datasets.

\paragraph{Annotation impact} Overall, without the copy mechanism, we can notice that even though annotations helped improve performances for non-pre-trained models, the impact is much more noticeable for pre-trained models which often reach much better performance with "tag-within" questions compared to the "raw-question" setting. 
Notably, we can report that non-pre-trained models gain around 2.02\% of F1 points on average going from no annotation ("raw-question") to "tag-within" question, whereas pre-trained models gain around 48.70\% of F1 points on average.






\subsection{Large Language Models}
We also conduct an empirical evaluation of two LLMs, namely Llama \cite{touvron2023llama} and Code Llama\cite{roziere2023code}. Due to computational and time limitations, we only exploit the LC-QuAD 2.0 dataset in these experiments due to its higher difficulty level. Our evaluation uses different portions of the training data, specifically 25\% and 50\% of the training set, corresponding to 5,440 and 10,880 entries from the train set, respectively. When fine-tuning with instruction, we also try using 100\% of the train set to get more insight into the models' behavior concerning the train sets' size. This is because this fine-tuning method shows more sensitivity to the train set size than the standard fine-tuning. 

\begin{figure}
\centering
    \includegraphics[scale=0.35]{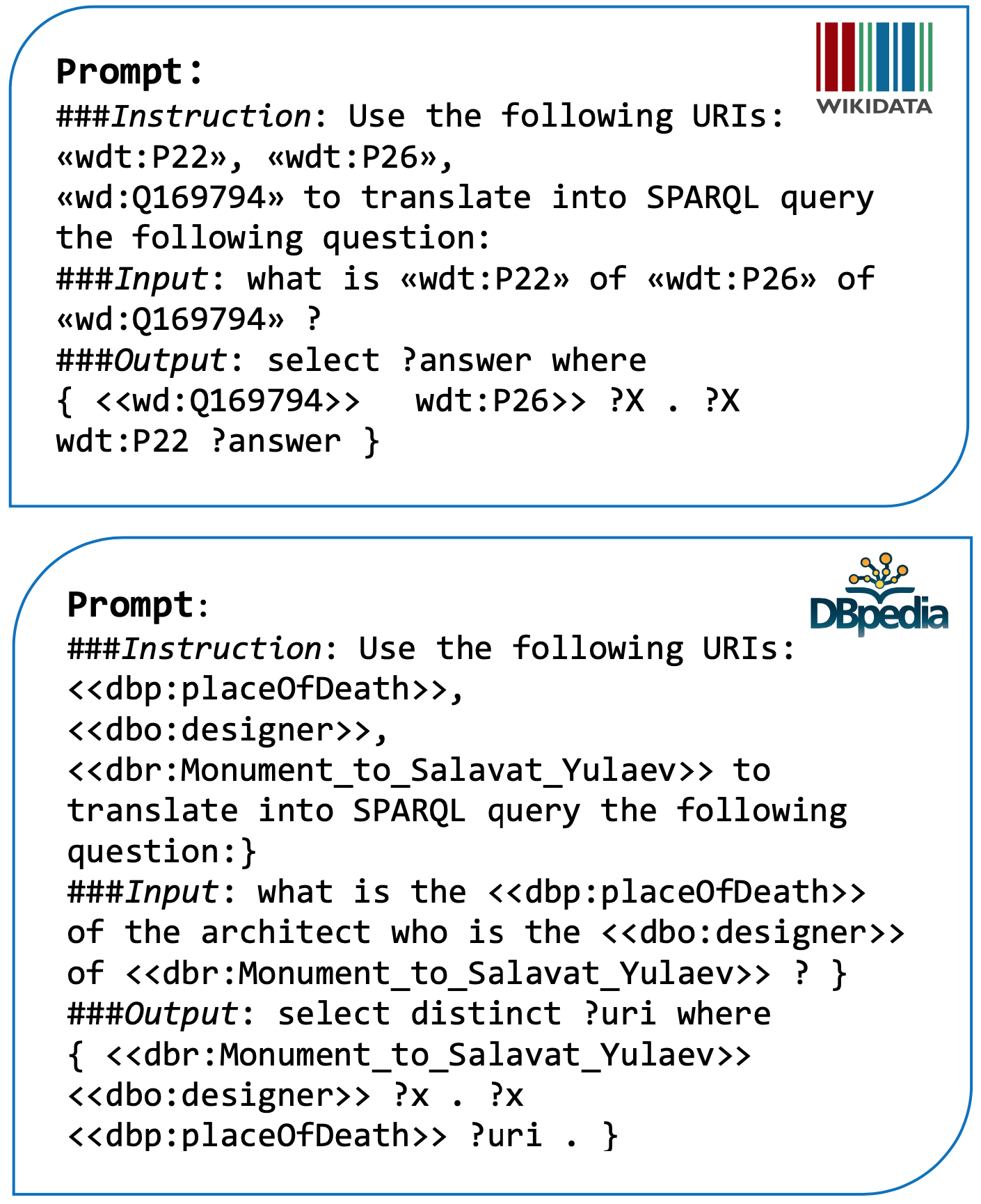}
    \caption{Example of prompts with the instruction, the input and the output.}
    \label{fig:prompt_examples}
\end{figure}

We explore two distinct fine-tuning approaches. First, the standard fine-tuning method entails providing the model with the question as input and the corresponding SPARQL query as output. Second, we adopt the instruction fine-tuning method, which augments the input-output pair with an additional instruction to guide the model's task comprehension. 
To simulate the "tag-end" instruction prompt, our instruction explicitly specifies the URIs to be used for generating the query, as illustrated in Figure \ref{fig:prompt_examples}. In contrast, the standard fine-tuning approach does not incorporate such explicit instructions; instead, it utilizes inputs with tags (questions wherein URIs corresponding to knowledge base elements are appended at the end of the input sequence) or without tags to emulate "raw-question".

As shown in Tables \ref{tab:llm-results1} and \ref{tab:llm-results2}, the performance of our Large Language Models (LLMs) falls short in comparison to our T5 and ConvSeq2Seq models. Both LLMs exhibit suboptimal performance levels, with the best F1 scores being 13\% and 23\%, respectively, for Llama and Code Llama. These results collectively indicate that, despite the tagging of questions and the extent of data utilized for fine-tuning, these LLMs do not substantially enhance their proficiency in generating effective SPARQL queries. It is important to note that the instruction boosts the performance for these models as it helps them better understand the task and mitigates undesirable behaviors such as generating extra text. 

\begin{table*}
\caption{Standard Fine Tuning of Large Language Models.}
\centering
\begin{threeparttable}
\begin{tabular}{c|c|ccc|ccc|ccc|ccc}
    \multicolumn{2}{c}{} & \multicolumn{12}{c}{} \\
    \cline{3-14}
    \multicolumn{2}{c}{} &  \multicolumn{3}{c}{Llama V2 7B (25\%)} \vline & \multicolumn{3}{c}{Llama V2 7B (50\%)} \vline & \multicolumn{3}{c}{ Code Llama V2 7B (25\%)} \vline & \multicolumn{3}{c}{Code Llama V2 7B (50\%)} \\
    \cline{3-14}
    \multicolumn{2}{c}{} & Bleu & Acc & F1 & Bleu & Acc & F1 & Bleu & Acc & F1 & Bleu & Acc & F1 \\
    \hline
    \multirow{2}{*}{LCQ 2}   & raw-question & 54 & 0.2 & 0.2 &  54 & 0.2 & 0.2 &  51 & 0.2 & 0.4 &  53 & 0.3 & 0.3 \\  
    & tag-end  & 55 & 0.2 & 0.3 & 56 & 0.3 & 0.3 & 54 & 0.2 & 0.1 & 56 & 0.3 & 0.3 \\ 
\hline
\end{tabular}
\label{tab:llm-results1}
\begin{tablenotes}
     \item[] All values in the table are in percentages.
  \end{tablenotes}
\end{threeparttable}
\end{table*}

\begin{table*}
\caption{Instruction Fine Tuning Large Language Models.}
\centering
\begin{threeparttable}
\begin{tabular}{c|c|ccc|ccc|ccc|ccc}
    \multicolumn{2}{c}{} & \multicolumn{12}{c}{} \\
    \cline{3-14}
    \multicolumn{2}{c}{} &  \multicolumn{3}{c}{Llama V2 7B (50\%)} \vline & \multicolumn{3}{c}{ LLama V2 7B (100\%)} \vline & \multicolumn{3}{c}{Code Llama V2 7B (50\%)} \vline & \multicolumn{3}{c}{Code Llama V2 7B (100\%)} \\
    \cline{3-14}
    \multicolumn{2}{c}{} & Bleu & Acc & F1 & Bleu & Acc & F1 & Bleu & Acc & F1 & Bleu & Acc & F1 \\
    \hline
    \multirow{2}{*}{LCQ 2}   & raw-question  & 54 & 13 & 13 &  52 & 12 & 12 &  54 & 11 & 12 & 43 &  21 & 20  \\  
     & tag-end  & 56 & 13 & 13 & 56 & 15 & 13 & 57 & 13 & 13 & 59 & 23 & 23 \\ 
    \hline 
\end{tabular}
\label{tab:llm-results2}
\begin{tablenotes}
     \item[] All values in the table are in percentages.
  \end{tablenotes}
\end{threeparttable}
\end{table*} 


\subsection{Best results} \label{sec:best-results}

We compare our accuracy scores with the former best approaches reported in \cite{copy-machanism} and \cite{banerjee2022modern} and our F1-scores with previous works as shown in Table \ref{tab:sota-models-comparison}. 
For each dataset, we report new state-of-the-art results 
in terms of F1 score and answer accuracy. 
The following elements summarize our best results for each dataset:
\begin{itemize}
    \item \textbf{LC-QuAD 1.0}: we report 96\% of F1 and 96\% of Accuracy with ConvSeq2Seq and 96\% of F1 and 97\% of Accuracy for T5, all with the copy mechanism and "tag-within" questions. This outperforms \cite{banerjee2022modern} by 5 points (of F1) and \cite{copy-machanism} by 2 points (of Accuracy).
    \item \textbf{LC-QuAD 2.0}: we report 94\% of F1 and 94\% of accuracy for T5 with "tag-within" questions. This outperforms \cite{banerjee2022modern} by 2 points of F1.
    \item \textbf{DBNQA}: we report 97\% of F1 and 97 of Accuracy with T5 for both "tag-within" and "tag-end" questions. This outperforms \cite{copy-machanism} by 10 points.
\end{itemize} 

However, we can note that while both studies \cite{copy-machanism, SPARQLParsing} use an answer-based score, there are some  methodological differences. On the one hand, \cite{copy-machanism} doesn't discard empty answers, which boost their performance. On the other hand, \cite{banerjee2022modern} uses the first non-empty answer among several beam search generated queries, which also might boost performances compared to our approach, which generates only one query. 

To compare our results to state-of-the-art models, we report the models with the best performance from the LC-QuAD 1.0 and LC-QuAD 2.0 leaderboard as shown in table \ref{tab:sota-models-comparison}. The first work, \cite{zhou2021improving}, focuses on multilingual question answering over knowledge graphs (KGQA) in a zero-shot transfer setting, utilizing unsupervised bilingual lexicon induction (BLI) to create augmented training data for multiple languages. 
The second work, \cite{purkayastha2022deep}, introduces SGPT, a novel approach combining end-to-end and modular systems with an advanced language model (GPT2\cite{radford2019language}). It employs an embedding technique for complex question patterns, incorporates graph-specific information into language model parameters, and introduces new metrics called SP-BLEU and SP-F1 metrics. 

\begin{table}[h!]
\caption{F1 Performance in \% for different SOTA models.}
\centering
\begin{threeparttable}
    \begin{tabular}{cccc}
        \toprule
        \textbf{SOTA Models} & LCQ 1 & LCQ 2 & DBNQA \\
        \midrule
        mBERT \cite{zhou2021improving} & 85 & - & - \\
        SGPT (Q,K) \cite{purkayastha2022deep} & - & 89 & - \\
        SubQG \cite{ding2019leveraging} & 85 & - & - \\
        \midrule
        \multicolumn{4}{c}{Results of \cite{banerjee2022modern}} \\
        \midrule
        PGN-BERT & 67 & 77 & - \\
        PGN-BERT-BERT & 88 & 86 & - \\
        BART & 84 & 64 & - \\
        T5-Small & 90 & 92 & - \\
        T5-Base & 91 & 91 & - \\
        \midrule
        \multicolumn{4}{c}{Our Models (Annotation: "tag-within")} \\
        \midrule
        T5  & \textbf{97} & \textbf{94} & \textbf{97} \\
        T5 Copy  & \textbf{96} & 72 & 56 \\
        BART Copy  & \textbf{96} & 85 & 91 \\
        ConvSeq2Seq Copy  & \textbf{96} & 69 & 90 \\
        Llama 2 7B & - & 16 & - \\
        Code Llama2 7B  & - & 31 & - \\
        \bottomrule
    \end{tabular}
    \begin{tablenotes}
     \item[]  Best performances are in bold.
    \end{tablenotes}
    \end{threeparttable}
    \label{tab:sota-models-comparison}
\end{table}
Given our RAM limitations, we employed the gradient accumulation method to maintain a valid comparison. By accumulating gradients four times for a batch size of 16, we effectively simulated a batch size of 64, in line with \cite{copy-machanism}. This approach yielded results consistent with those depicted in table \ref{tab:gradient_accumulation_impact}. Here we report only the BLEU and Accuracy since those are the metrics used by \cite{copy-machanism}. We observe slightly lower performance in LC-QuAD 1.0. 

\begin{table}[h!]
\caption{BLEU performance (\%) comparison with\cite{copy-machanism}}
\centering
\begin{threeparttable}
\resizebox{\linewidth}{!}{
\begin{tabular}{c|c|cc|cc|cc|cc}
\multicolumn{2}{c}{} & \multicolumn{4}{c}{Transformer} \vline & \multicolumn{4}{c}{ConvSeq2Seq} \\
\cline{3-10}
\multicolumn{2}{c}{} &  \multicolumn{2}{c}{Base} \vline & \multicolumn{2}{c}{Copy} \vline &  \multicolumn{2}{c}{Base} \vline & \multicolumn{2}{c}{Copy}\\
\cline{3-10}
\multicolumn{2}{c}{} & Bleu & Acc & Bleu & Acc & Bleu & Acc & Bleu & Acc \\
\hline
\multirow{1}{*}{LCQ 1 } &  Reform. Qst    & 49 \textcolor{blue}{43} & 32 \textcolor{blue}{28} & / & / 
& 49 \textcolor{blue}{42} & 40 \textcolor{blue}{39} & / & / \\ 
\hline
\multirow{3}{*}{DBNQA} &  "raw-question"    & 64 \textcolor{blue}{58} & 46 \textcolor{blue}{47} & / & /  
         & 67 \textcolor{blue}{71} & 45 \textcolor{blue}{52} & / & / \\       
& tag within & 65 \textcolor{blue}{58} & 47 \textcolor{blue}{45} & 93 \textcolor{blue}{95} & 85 \textcolor{blue}{86} 
         & 67 \textcolor{blue}{75} & 45 \textcolor{blue}{59} & 95 \textcolor{blue}{97} & 86 \textcolor{blue}{92} \\  
        \hline
        \end{tabular}}
        \begin{tablenotes}
     \item[]The results in blue are obtained using Gradient Accumulation. \\ Reform. Qst: natural reformulation of question done by an expert.
  \end{tablenotes}
\end{threeparttable}
\label{tab:gradient_accumulation_impact}
\end{table}

\section{Error Analysis} \label{sec:models-comparison-error-analysis}
To further study the differences between all our models, we perform an error analysis to identify the  kind of errors made by the models. 

\subsection{Token/Error Types}
The tokens comprising the SPARQL queries generated by our models can be systematically categorized into six distinct classes: URIs, SPARQL keywords, functions or operators, literals, variables, and unknown tokens. Consequently, our analysis identifies six different error types, each attributed to the specific token category where it was observed. Therefore, we delineate the following error categories:

\begin{itemize}
    \item Knowledge graph URIs (abbreviated as \textbf{URIs}): indicates errors due to a difference in URIs between the reference and the prediction. We use regular expressions that detect prefixes (complete or truncated) and the identifier of each entity or resource (e.g., wdt:P31, http://www.wikidata.org/entity/Q146). "Fake URIs" are tokens with a URI pattern that do not exist in 
    knowledge base.  
    \item SPARQL keywords (abbreviated as \textbf{SVocab})): These include errors due to an incorrect generation of a SPARQL keyword at a specific position in the query. Some of the SPARQL keywords considered are, for instance, "select", "ask", or "describe", "distinct", "reduced", "where", "limit".
    \item Functions or operators (abbreviated as \textbf{Fct}): These include errors due to reference and prediction mismatch on functions and operators. Some examples of SPARQL functions considered are: "=", "!=", "<", ">", "<=", ">=", "+", "-", "*", "/", "str", "ucase", "lcase" and "concat".
    \item Literals (abbreviated as \textbf{Lit}): include faulty generations of a literal value. A literal is recognized as a string in double quotes or can be a numeric literal (ex 42), a boolean literal (true, false), or a date.
    \item Variables (abbreviated as \textbf{Var}): Errors made on variables. 
    \item Unknown (abbreviated by \textbf{Unk}) designates an error made on the generation of the token "<unk>" that represents all the OOV elements.
\end{itemize}

\subsection{Error Type Distribution}
We align the reference query and the  generated query and detect mismatches at the token level.

Based on token types, our  objective is to determine which type of token is generated instead of 
the expected type. 
This makes it possible to study exactly how models are wrong: for example, how often does a model generate a URI instead of a SPARQL keyword, how often does it misplace a URI in triples (when there is more than one URI in the same query). We also measure if URIs in particular can be hallucinated, that is, how often does a query include fake URIs. Examples of errors are shown in Table \ref{tab:sparql-queries-errors}. \\

\begin{table*}
\caption{Some examples of errors on SPARQL queries.}
\centering
\begin{tabularx}{\textwidth}{lXX}
\toprule
\textbf{Errors Types} & \textbf{Target Query} & \textbf{Predicted Query} \\
\midrule
\textbf{URIs} & \texttt{select distinct ?uri where \{ \textbf{dbr:Philip\_Novak }dbp:mainInterests ?uri \}} & \texttt{select distinct ?uri where \{ \textbf{dbr:Timothy\_Morton} dbp:mainInterests ?uri \}} \\
\textbf{"Fake URIs"} & \texttt{select distinct ?obj where \{ wd:Q206856 wdt:P166 ?obj . ?obj \textbf{wdt:P31} wd:Q131647 \}} & \texttt{select distinct ?obj where \{ wd:Q206856 wdt:P166 ?obj . ?obj \textbf{w:Pq31} wd:Q131647 \}} \\

\textbf{SVocab} & \texttt{select distinct ?uri \textbf{where} \{ dbr:Dan\_Otero dbo:debutTeam ?uri \}} & \texttt{select distinct ?uri \textbf{\{} dbr:John\_Estes dbo:debutTeam ?uri . ?x dbo:debutTeam ?uri . \}} \\
\textbf{Fonction} & \texttt{ask where wd:Q2084454 wdt:P5066 ?obj \textbf{filter}(?obj = 22.4) } & \texttt{ask where wd:Q2084454 wdt:P5066 ?obj \textbf{where}(?obj = 22\_4) } \\
\textbf{Literal} & \texttt{ask where wd:Q2084454 wdt:P5066 ?obj filter(?obj = \textbf{22.4}) } & \texttt{ask where wd:Q2084454 wdt:P5066 ?obj filter(?obj = \textbf{22\_4}) } \\
\bottomrule
\end{tabularx}
\label{tab:sparql-queries-errors}
\end{table*}

The type errors are detailed in Table \ref{tab:sparql-queries-errors}.
\paragraph{Error distribution for non-pretrained models}
For the LC-QuAD 1.0 dataset, the base non-pre-trained models ConvSeq2Seq and Transformer 
have a lot of difficulties generating the correct queries because they make errors practically on all types such as URIs, SVocab, and the Unknowns. This confusion of models is further accentuated on the LC-QuAD 2.0 dataset in which we also observe errors at the Variable level. However, in copy-based models, the errors are only at the level of URIs and SPARQL Vocabulary for all datasets and are due to the incorrect ordering of tokens 
for these two types during generation. Indeed, copying helps to respect the template of the reference query but does not help to choose the right URIs among the candidates for copying. With the "tag-end" setting, we note a lot of errors on the URIs, and the rate of Fake-URIs is slightly higher compared to "tag-within" and "raw-question", which are settings that lead to lower error percentages. 

\paragraph{Error distribution for pretrained models}
We have roughly the same observations with the pre-trained models except that there are unknown tokens errors in LC-QuAD 2.0, which has a much higher number of OOV tokens in the test set, as shown in Table \ref{tab:voc}. Indeed, on LC-QuAD 1.0, BART and T5, with or without the copy mechanism, make errors in the generation of URIs and SVocab. Copy-based models make wrong choices of URIs among the copy URI candidates. As previously with NPLMs, on LC-QuAD 2.0, some errors are made on variables due to the greater complexity of this dataset. 
From the point of view of the impact of the annotation, we see that without the copy, we have approximately the same rate of URI errors (41.7 \%) whatever the tagging method, and this rate increases by 12 \% in the "raw-question" data. 




\paragraph{Error distribution for large pretrained models}
As for the LLMs, the generation errors are essentially made very largely at the level of the URIs, then in a smaller percentage on Variables, and finally in a small percentage on the SVocab regardless of the fine-tuning method used. Almost at all positions where URIs are expected, LLMs generate incorrect URIs and occasionally predict URIs that do not exist. This hallucination is partly due to the absence of the copying mechanism forcing the models to trust their pre-training knowledge. In addition, the Variables and SVocab are mixed together in the query, i.e., at the position where a variable is expected, an SVocab is generated, and vice versa. 

\paragraph{Summary.}
The error distribution in query generation varies among different types of models but is mostly made on the URIs and SVocab levels. NPLMs, when generating queries for LC-QuAD 1.0 and LC-QuAD 2.0 datasets, struggle with errors across various token types, including URIs, SVocab, and Unk. Copying helps maintain query structure but doesn't always select the correct URIs. PLMs have a similar pattern of errors, with less Unk tokens due to their larger vocabulary. These models also exhibit URIs and SVocab generation errors, especially on LC-QuAD 1.0. With LLMs, most errors occur with URIs, followed by Variables and SVocab, and this hallucination is exacerbated by the absence of a copy mechanism. Variables and SVocab tokens are often interchanged during generation. In all cases, maintaining tagging methods is beneficial, and the "tag-within" strategy works best when using the copy mechanism. 

\subsection{Generalization capabilities of the best models}
\label{para:generalization-capabilities}
For this part, we considered the best models and optimal configurations in two datasets and evaluated their generalization capacity. For both LC-QuAD 1.0 and LC-QuAD 2.0, the best models are T5 Base and ConvSeq2Seq Copy, with the "tag-within" setting. To test the generalization abilities of these models, we carried out the following three experiments:
\begin{enumerate}
     \item We first trained these models with the original (template-based) questions and then tested them on the test set's reformulated questions. Our objective here is to test the ability of the models to handle natural questions when trained on template-based questions.
     \item We trained these models with the original questions and then tested them on the train set's reformulated questions. Our objective here is to test the ability of the models to handle natural questions that are paraphrases of template-based questions encountered during the training phase.
     \item We trained the models on the train set 's reformulated questions and tested them on the test set's reformulated questions. Our objective here is to measure the performance of the models on natural, non-template-based questions.
\end{enumerate}

The results of these experiments are shown in Table \ref{tab:TrainOQ-TestRQ-Of-TestSet}, Table \ref{tab:TrainOQ-TestRQ-Of-TrainSet} and Table \ref{tab:TrainRQ-TestRQ}. Since copying is performed on tagged questions, there aren't any results for the ConvSeq2Seq Copy models in the "raw-question" configuration. The "tag-within" method is not shown in these tables, as reformulations are exclusively applied to "tag-end" questions. This is because when we use reformulated questions, we can no longer leverage any template for tagging the question. 

\begin{table*}
\caption{Performances after training on \textbf{original questions} and testing on the \textbf{test set reformulated questions}}
\centering
\footnotesize
\begin{tabular}{cccccc|cccccc}
\toprule
\multicolumn{12}{c}{\textbf{LC-QuAD 1.0}} \\
\midrule
\multicolumn{2}{c}{\multirow{2}{*}{\textbf{Models}}} &  & \multicolumn{3}{c}{\textbf{Metrics}} & \multicolumn{6}{c}{\textbf{Error Distribution}} \\
\cmidrule(lr){4-12}
\multicolumn{2}{c}{} & \textbf{Tagging} & Bleu & Acc & F1 & Uri & SVoc & Lit & Fct & Var & Unk \\
\midrule
\multirow{2}{*}{PLM}

 & \multirow{2}{*}{T5 Small} 
 & "raw-question" & 47 & 23 & 22 & 39 & 37 & 3 & 2 & 19 & 0  \\ &
 & "tag-end" & 64 & 34 & 34 & 47 & 53 & 0 & 0 & 0 & 0  \\ 
\midrule
\multirow{1}{*}{NPLM}
& \multirow{1}{*}{ConvSeq2Seq Copy} 
& "tag-end" & 55 & 23 & 24 & 40 & 60 & 0 & 0 & 0 & 0  \\ 
\midrule
\multicolumn{12}{c}{\textbf{LC-QuAD 2.0}} \\
\midrule
\multicolumn{2}{c}{\multirow{2}{*}{\textbf{Models}}} &  & \multicolumn{3}{c}{\textbf{Metrics}} & \multicolumn{6}{c}{\textbf{Error Distribution}} \\
\cmidrule(lr){4-12}
\multicolumn{2}{c}{} & \textbf{Tagging} & Bleu & Acc & F1 & Uri & SVoc & Lit & Fct & Var & Unk \\
\midrule
\multirow{2}{*}{PLM}
& \multirow{2}{*}{T5 Small} 
& "raw-question" & 25 & 0.3 & 0.2 & 32 & 34 & 3 & 2 & 29 & 0  \\ &
& "tag-end" & 51 & 32 & 32 & 25 & 39 & 2 & 3 & 31 & 0  \\ 
\midrule
\multirow{1}{*}{NPLM}
& \multirow{1}{*}{ConvSeq2Seq Copy} 
& "tag-end" & 50 & 11 & 11 & 66 & 13 & 2 & 0 & 19 & 0  \\ 
\bottomrule
\end{tabular}
\label{tab:TrainOQ-TestRQ-Of-TestSet}
\end{table*}
    
\begin{table*}
\caption{Performances after training on \textbf{original questions} and testing on the \textbf{train set reformulated questions}}
\centering
\begin{tabular}{cccccc|cccccc}
\toprule
\multicolumn{12}{c}{\textbf{LC-QuAD 1.0}} \\
\midrule
\multicolumn{2}{c}{\multirow{2}{*}{\textbf{Models}}} &  & \multicolumn{3}{c}{\textbf{Metrics}} & \multicolumn{6}{c}{\textbf{Error Distribution}} \\
\cmidrule(lr){4-12}
\multicolumn{2}{c}{} & \textbf{Tagging} & Bleu & Acc & F1 & Uri & SVoc & Lit & Fct & Var & Unk \\
\midrule
\multirow{2}{*}{PLM}
& \multirow{2}{*}{T5 Small} 
& "raw-question" & 44 & 25 & 25 & 34 & 35 & 8 & 5 & 18 & 0  \\ &
& "tag-end" & 65 & 34 & 36 & 47 & 53 & 0 & 0 & 0 & 0  \\ 
\midrule
\multirow{1}{*}{NPLM}
& \multirow{1}{*}{ConvSeq2Seq Copy} 
& "tag-end" & 62 & 32 & 32 & 25 & 64 & 11 & 0 & 0 & 0  \\ 
\midrule
\multicolumn{12}{c}{\textbf{LC-QuAD 2.0}} \\
\midrule
\multicolumn{2}{c}{\multirow{2}{*}{\textbf{Models}}} &  & \multicolumn{3}{c}{\textbf{Metrics}} & \multicolumn{6}{c}{\textbf{Error Distribution}} \\
\cmidrule(lr){4-12}
\multicolumn{2}{c}{} & \textbf{Tagging} & Bleu & Acc & F1 & Uri & SVoc & Lit & Fct & Var & Unk \\
\midrule
\multirow{2}{*}{PLM}
& \multirow{2}{*}{T5 Small} 
& "raw-question" & 23 & 0.1 & 0.1 & 22 & 36 & 8 & 5 & 29 & 0  \\ 
&  & "tag-end" & 46 & 24 & 24 & 20 & 40 & 6 & 3 & 31 & 0  \\ 
\midrule
\multirow{1}{*}{NPLM}
& \multirow{1}{*}{ConvSeq2Seq Copy} 
& "tag-end" & 52 & 15 & 15 & 65 & 14 & 0 & 0 & 21 & 0  \\ 
\bottomrule
\end{tabular}
\label{tab:TrainOQ-TestRQ-Of-TrainSet}
\end{table*}

\begin{table*}
\caption{Performances after training on \textbf{reformulated questions} and testing on the \textbf{test set reformulated questions}}
\centering
\footnotesize
\begin{tabular}{cccccc|cccccc}
\toprule
\multicolumn{12}{c}{\textbf{LC-QuAD 1.0}} \\
\midrule
\multicolumn{2}{c}{\multirow{2}{*}{\textbf{Models}}} &  & \multicolumn{3}{c}{\textbf{Metrics}} & \multicolumn{6}{c}{\textbf{Error Distribution}} \\
\cmidrule(lr){4-12}
\multicolumn{2}{c}{} & \textbf{Tagging} & Bleu & Acc & F1 & Uri & SVoc & Lit & Fct & Var & Unk \\
\midrule
\multirow{2}{*}{PLM}
 & \multirow{2}{*}{T5 Small} 
& "raw-question" & 77 & 68 & 69 & 44 & 34 & 5 & 1 & 16 & 0  \\ &
& "tag-end" & 88 & 72 & 73 & 61 & 39 & 0 & 0 & 0 & 0  \\ 
 \midrule
 \multirow{1}{*}{NPLM}
 & \multirow{1}{*}{ConvSeq2Seq Copy} 
& "tag-end" & 78 & 52 & 54 & 63 & 36 & 0 & 0 & 0 & 1  \\ 
\midrule
\multicolumn{12}{c}{\textbf{LC-QuAD 2.0}} \\
\midrule
\multicolumn{2}{c}{\multirow{2}{*}{\textbf{Models}}} &  & \multicolumn{3}{c}{\textbf{Metrics}} & \multicolumn{6}{c}{\textbf{Error Distribution}} \\
\cmidrule(lr){4-12}
\multicolumn{2}{c}{} & \textbf{Tagging} & Bleu & Acc & F1 & Uri & SVoc & Lit & Fct & Var & Unk \\
\midrule
\multirow{2}{*}{PLM}
& \multirow{2}{*}{T5 Small} 
& "raw-question" & 63 & 2.1 & 2.1 & 56 & 20 & 2 & 1 & 21 & 0  \\ &  
& "tag-end" & 85 & 76 & 75 & 41 & 15 & 3 & 1 & 40 & 0  \\ 
\midrule
\multirow{1}{*}{NPLM}
& \multirow{1}{*}{ConvSeq2Seq Copy} 
& "tag-end" & 56 & 25 & 26 & 56 & 20 & 0 & 0 & 24 & 0  \\ 
\bottomrule
\end{tabular}
\label{tab:TrainRQ-TestRQ}
\end{table*}

It is clear from Table \ref{tab:TrainOQ-TestRQ-Of-TestSet} that query generation is more challenging, as the train questions' structures differ from those of the test set. Conversely, as shown in Table \ref{tab:TrainOQ-TestRQ-Of-TrainSet}, the models exhibit approximately the same performance in query generation when reformulated questions are paraphrases of train questions. 
Nevertheless, a substantial decrease in model performance is observed when models trained on original questions are tested with reformulated questions. Furthermore, a significant decline is observed across all configurations for models trained and tested on reformulated questions, as shown in Table \ref{tab:TrainRQ-TestRQ}. This reaffirms the notion that templates constitute a significant component for achieving a robust alignment between a question and its corresponding SPARQL query. But this also highlights that current models are not yet ready to be used - as is - for natural questions.  It is worth noting that T5 Base is less affected by the "noise" introduced with question reformulation compared to ConvSeq2Seq, owing to its pre-training, which endows it with language knowledge and enhanced adaptability to changes while preserving semantics (e.g., synonyms).

\section{Discussion} \label{sec:discussion}
Overall, the PLMs outperform the NPLMs and the LLMs in both LC-QuAD 1.0 and LC-QuAD 2.0, even though we obtained pretty good results with the ConvSeq2Seq model with copy. 
Counter-intuitively, we obtained low performance with the LLMs after trying various kinds of training (standard fine-tuning, instruction tuning) with different subsets of the train set. This low performance is partly justified by the small amount of data related to SPARQ in the datasets used to pretrain these models. For example, in Code Llama\cite{roziere2023code}, most data relates to Python, C++, Java, PHP, TS, C\#, and Bash. 

\subsection{The impact of the tagging and copy mechanisms}

Question annotations always positively impact the performance of the generation for non-pretrained and pretrained language models. From Table \ref{tab:errAnal1} - \ref{tab:errAnal4}, we observe that in less complex datasets like LC-QuAD 1.0 and DBNQA, the results are similar accross models, even though T5 seems to be consistently the top performer. With LC-QuAD 2.0, the PLMs outperformed the other models with a large margin. However, the pretraining in T5 does not solve the problem of identifying the correct URIs, which is why question annotation is important. LLMs obtained a lower performance for SPARQL query generation, even when pre-trained on code generation. The importance of question annotation for improved performance shows the necessity of having better question's semantic representations adaptable to knowledge base schema and resources. In future work, we plan to investigate how to learn objective functions that target annotation and generation at the same time. 

\paragraph{Difficulty of the "tag-end" setting for the copy mechanism} 
We noticed that many models struggle with the "tag-end" questions when using the copy mechanism, this mostly occurs with some Transformer-based models, namely our non-pre-trained Transformer and T5. This is probably due to the fact that placing KB elements in the query is conditioned by the adequate location of the corresponding URI in the question. In the "tag-within" setting, the cross attention heads easily learn to map the position in the question to the right position in the query. However, with "tag-end" questions, the model has to associate the KB element to its label and then map this label to the corresponding natural language mention in the original question. These additional steps might be the source of the challenge observed with some Transformer models. We can suppose that the convolution operation is more suited to addressing these steps, maybe because of the proximity between the URI and its label at the end of the annotated questions. Additionally, BART also seems to overcome these difficulties. We can suggest that its pre-training tasks based on denoising might have led to better short context associations. 

\paragraph{Limitations of the copy mechanism}
The main limitation of this mechanism is that it mainly relies on the structure of the question-query pairs and on the positional mapping of the URIs between questions and queries. 
It is possible that these models only focus on the task of question template - query template mapping and placeholder filling. 
This might pose problems when we expose the model to a question-query pair that is generated by a template not associated to enough or any training entries, or, as we saw, to reformulated questions that do not follow a known template. 

Another limitation of our copy-based architectures is that they are blind to the URIs' semantics. The encoder-decoder does not see these tokens since they are masked, and the copy mechanism only uses positions to copy the tokens. This might cause problems when the expected query is independent of the question structure. When we mask the URIs, the copy layer calculates the probability distributions of the elements to copy from a list of candidate URIs that occur in the query. This has the effect of causing errors at the level of the URIs because, as we have seen in the analysis of the errors, we have an essential proportion of mistakes due to the wrong choice of URIs to copy among the candidates. For example, from Table \ref{tab:models-comparison}, we can see that 27\% of the incorrect queries generated by the BART model with copy are caused by a wrong choice among the candidate URIs. By considering the masking that comes with the copy, our hypothesis is that we increase the difficulty to elect a candidate URI for the copy. This masking also accentuates the tendency of the models to choose the most frequent query template for questions having the same question template. 

These limitations could be addressed by unmasking URIs in the case of pre-trained models. 
Yet we would be facing a "spelling" problem that is the problem that occurs when the tokenizer split the URIs into fragments and does not properly merge them back at generation. Other types of copy mechanisms could be explored that would not require masking the URIs, would keep them as a single token without adding them to the model's vocabulary. This will be investigated in future work.  

\subsection{Error types}
\paragraph{Comparison of models in terms of performance and distribution of the error types}
Table \ref{tab:models-comparison} lists the overall performance of our different models for all configurations and the distribution of error types in percentage. To fully understand this table, the reader first needs to look at the metrics that give the overall performance (BLUE score, Accuracy, or F1 score) before looking at the distribution of errors because a configuration can have a high error rate for a given error type while conserving a good performance. All values in the "Error Distribution" columns represent errors in query generation on the corresponding type. For example, in the first line of Table \ref{tab:models-comparison}, we see that the BART model has a BLEU, Accuracy, and F1 of 84\%, 72\%, and 75\%, respectively, and that 67\% of incorrect queries are due to incorrect URIs with 52\% being URIs that exist in the knowledge base that are incorrectly placed in the SPARQL query and 15\% being "Fake URIs." 

Considering both LC-QuAD 1.0 and LC-QuAD 2.0 with the best annotation method, which is "tag-within," we obtain an average of 28\% of incorrect URIs and 4\% of "Fake URIs" for the T5 model without the copy mechanism.
Conversely, for the model ConvSeq2Seq, we have 27.5\% of incorrect URIs and 0\% of "Fake URIs." Plus, there is 1\% more "Fake Uris" for the "tag-end" setting compared to the "tag-within" setting.


    %

\begin{table*}
\caption{Models Comparison in terms of performance (based on BLEU, Accuracy and F1-score) and error type distribution.}
\centering
\begin{threeparttable}
\begin{tabular}{cccccc|ccccccc}
\toprule
\multicolumn{12}{c}{\textbf{LC-QuAD 1.0}} \\
\midrule
\multicolumn{2}{c}{\multirow{2}{*}{\textbf{Models}}} &  & \multicolumn{3}{c}{\textbf{Metrics (\%)}} & \multicolumn{7}{c}{\textbf{Error Distribution* (\%)}} \\
\cmidrule(lr){4-13}
\multicolumn{2}{c}{} & \textbf{Tagging} & Bleu & Acc & F1 & Uris & FakeUris & SVocab & Lit & Fct & Var & Unk \\
\midrule
\multirow{12}{*}{PLM}
 & \multirow{3}{*}{Bart} & raw-question & 84 & 72 & 75 & 52 & 15 & 33 & 0 & 0 & 0 & 0  \\ 
 &  & tag-within & 96 & 87 & 87 & 45 & 9 & 46 & 0 & 0 & 0 & 0  \\
 &  & tag-end & 94 & 84 & 85 & 41 & 13 & 46 & 0 & 0 & 0 & 0  \\ 
\cmidrule(lr){2-13}
 & \multirow{3}{*}{Bart Copy} & raw-question & - & - & - & - & - & - & - & - & - & -  \\ 
 &  & tag-within & 99 & 96 & 96 & 27 & 0 & 73 & 0 & 0 & 0 & 0  \\
 &  & tag-end & 97 & 94 & 93 & 37 & 0 & 63 & 0 & 0 & 0 & 0  \\ 
\cmidrule(lr){2-13}
 & \multirow{3}{*}{T5 Small} & raw-question & 81 & 54 & 57 & 58 & 15 & 26 & 0 & 0 & 0 & 1  \\ 
 &  & tag-within & 99 & 97 & 97 & 25 & 8 & 67 & 0 & 0 & 0 & 0  \\ 
 &  & tag-end & 98 & 96 & 96 & 32 & 11 & 57 & 0 & 0 & 0 & 0  \\ 
\cmidrule(lr){2-13}
 & \multirow{3}{*}{T5 Copy} & raw-question & - & - & - & - & - & - & - & - & - & -  \\ 
 &  & tag-within & 99 & 96 & 96 & 23 & 0 & 76 & 0 & 1 & 0 & 0  \\
 &  & tag-end & 98 & 92 & 93 & 46 & 0 & 65 & 0 & 0 & 0 & 1  \\ 
\midrule
 \multirow{12}{*}{NPLM}
  & \multirow{3}{*}{Transformer} & raw-question & 69 & 33 & 37 & 38 & 9 & 19 & 0 & 0 & 0 & 34  \\ 
 &  & tag-within & 82 & 41 & 44 & 17 & 0 & 75 & 0 & 0 & 0 & 8  \\ 
 &  & tag-end & 74 & 38 & 41 & 48 & 18 & 14 & 0 & 0 & 0 & 20  \\ 
\cmidrule(lr){2-13}
 & \multirow{3}{*}{Transformer Copy} & raw-question & - & - & - & - & - & - & - & - & - & -  \\ 
 &  & tag-within & 99 & 95 & 95 & 16 & 0 & 76 & 0 & 0 & 0 & 8  \\ 
 &  & tag-end & 49 & 1.8 & 1.8 & 85 & 0 & 15 & 0 & 0 & 0 & 0  \\ 
\cmidrule(lr){2-13}
& \multirow{3}{*}{ConvSeq2Seq} & raw-question & 69 & 23 & 27 & 37 & 12 & 14 & 0 & 0 & 0 & 37  \\ 
 &  & tag-within & 78 & 28 & 32 & 27 & 7 & 9 & 0 & 0 & 0 & 57  \\ 
 &  & tag-end & 75 & 25 & 28 & 55 & 9 & 18 & 0 & 0 & 0 & 18  \\ 
\cmidrule(lr){2-13}
 & \multirow{3}{*}{ConvSeq2Seq Copy} & raw-question & - & - & - & - & - & - & - & - & - & -  \\ 
 &  & tag-within & 99 & 96 & 96 & 14 & 0 & 76 & 0 & 1 & 0 & 9  \\ 
 &  & tag-end & 90 & 77 & 78 & 81 & 0 & 18 & 0 & 0 & 0 & 1  \\ 
\midrule
\multicolumn{12}{c}{\textbf{LC-QuAD 2.0}} \\
\midrule
\multirow{12}{*}{PLM}
 & \multirow{3}{*}{Bart} & raw-question & 72 & 4.7 & 4.8 & 51 & 3 & 31 & 2 & 3 & 10 & 0  \\ 
 &  & tag-within & 87 & 84 & 84 & 14 & 1 & 59 & 7 & 5 & 14 & 0  \\
 &  & tag-end & 88 & 84 & 84 & 65 & 2 & 7 & 2 & 6 & 18 & 0  \\ 
\cmidrule(lr){2-13}
 & \multirow{3}{*}{Bart Copy} & raw-question & - & - & - & - & - & - & - & - & - & -  \\ 
 &  & tag-within & 95 & 85 & 85 & 58 & 0 & 1 & 9 & 0 & 32 & 0  \\
 &  & tag-end & 89 & 72 & 72 & 42 & 0 & 27 & 1 & 2 & 28 & 0  \\
  
\cmidrule(lr){2-13}
& \multirow{3}{*}{T5 Small} & raw-question & 79 & 13 & 14 & 91 & 3 & 0 & 0 & 0 & 6 & 0  \\ 
 &  & tag-within & 95 & 94 & 94 & 31 & 0 & 12 & 18 & 1 & 38 & 0  \\ 
 &  & tag-end & 90 & 90 & 90 & 21 & 1 & 1 & 0 & 0 & 77 & 0  \\ 
\cmidrule(lr){2-13}
 & \multirow{3}{*}{T5 Copy} & raw-question & - & - & - & - & - & - & - & - & - & -  \\ 
 &  & tag-within & 90 & 72 & 72 & 38 & 0 & 31 & 1 & 4 & 26 & 0  \\
 &  & tag-end & 87 & 58 & 58 & 29 & 0 & 38 & 3 & 7 & 23 & 0  \\ 
\midrule
 \multirow{12}{*}{NPLM}
  & \multirow{3}{*}{Transformer} & raw-question & 57 & 0.8 & 1.1 & 30 & 2 &23 & 3 & 2 & 20 & 20  \\ 
 &  & tag-within & 59 & 1.5 & 1.6 & 29 & 1 & 26 & 3 & 2 & 18 & 21  \\ 
 &  & tag-end & 66 & 1.4 & 1.7 & 41 & 1 & 12 & 2 & 1 & 15 & 28  \\ 
\cmidrule(lr){2-13}
 & \multirow{3}{*}{Transformer Copy} & raw-question & - & - & - & - & - & - & - & - & - & -  \\ 
 &  & tag-within & 82 & 69 & 69 & 37 & 0 & 23 & 0 & 1 & 39 & 0  \\ 
 &  & tag-end & 57 & 2.2 & 2.2 & 68 & 0 & 11 & 7 & 0 & 14 & 0  \\ 
\cmidrule(lr){2-13}
& \multirow{3}{*}{ConvSeq2Seq} & raw-question & 76 & 9.8 & 11 & 19 & 14 & 5 & 2 & 0 & 9 & 51  \\ 
 &  & tag-within & 76 & 8.8 & 10 & 25 & 11 & 2 & 1 & 0 & 4 & 57  \\ 
 &  & tag-end & 78 & 14 & 15 & 19 & 18 & 0 & 2 & 0 & 8 & 53  \\ 
\cmidrule(lr){2-13}
 & \multirow{3}{*}{ConvSeq2Seq Copy} & raw-question & - & - & - & - & - & - & - & - & - & -  \\ 
 &  & tag-within & 89 & 69 & 69 & 41 & 0 & 28 & 1 & 2 & 27 & 1  \\ 
 &  & tag-end & 90 & 66 & 66 & 76 & 0 & 3 & 2 & 0 & 19 & 0  \\ 

 \midrule
 \multirow{2}{*}{LLM (100\% of Train set)}
& \multirow{1}{*}{Llama V2 7B}     & Instruct FT & 56 & 15 & 13 & 50 & 10 & 25 & 5 & 2 & 8 & 0  \\ 
\cmidrule(lr){2-13}
& \multirow{1}{*}{Code LlamaV2 7B} & Instruct FT & 59 & 23 & 23 & 58 & 3 & 26 & 5 & 2 & 6 & 0  \\ 
\bottomrule
\end{tabular}
  \begin{tablenotes}
   \item[*]  It is essential to look at the general performance before looking at the error distribution to evaluate the rates in relative values properly. The All the values in this table are in percentage.
  \end{tablenotes}
\end{threeparttable}
\label{tab:models-comparison}
\end{table*}

To go deeper into the analysis of the errors for the best models (T5 and ConvSeq2Seq Copy), we also include Tables \ref{tab:errAnal1} - \ref{tab:errAnal4} which give an analysis of the errors by showing which token types are generated in place of the expected ones. These analyses reveal that often, instead of the expected URIs,  incorrect URIs are predicted by all models. 
While copy models go wrong by making wrong choices between candidates, base models occasionally generate URIs that do not exist, as we can see in Table \ref{tab:models-comparison}. There are more "Fake URIs" with LC-QuAD 1.0 than LC-QuAD 2.0 because the morphology of URIs in DBpedia is more sensitive to errors. On the other hand, with Wikidata, an error can lead to another URI, which, even if not the expected one, exists in the knowledge base. Finally, we find a relatively high frequency of SPARQL vocabulary (SVocab) or Variables that are put in place of URIs.

\subsection{The generalization capabilities of the models}
As shown in section \ref{para:generalization-capabilities}, after conducting various experiments to test generalization capabilities, including training on original questions and testing on reformulated questions, results revealed that models face challenges when the question structure differs during testing, resulting in a critical decreased performance. Models trained and tested on reformulated questions also exhibited lower performance, highlighting the importance of templates for aligning questions with SPARQL queries. Considering the best models for the experiments on original questions with "tag-end" setting, the average F1 score is respectively 93\% and 72\% for T5 and ConvSeq2Seq Copy, respectively. On the other hand, for reformulated questions, the average F1 score is 33\% and 17.5\%, respectively for T5 and ConvSeq2Seq Copy. Thus, we observe an overall drop in performance of 60\% for T5 and a drop of 54.5\% for NPLMs compared to the results with original questions. Pre-trained Language Models are more resilient to question reformulation than Non-Pre-trained Language Models. 



\subsection{Other considerations}
\paragraph{LC-QuAD 2.0.} Most results on LC-QuAD 2.0 are low (compared to results on other datasets) and in particular with non-pre-trained models. This is most probably because of the large number of unknown words in the test set (5,751 in LC-QuAD 2.0 versus 368 in LC-QuAD 1.0 as shown in Table \ref{tab:voc}) and the URIs formulation in Wikidata which is a set of numbers that cannot be mapped to known semantic structures 
compared to DBPedia URIs which use words, 
and to a lesser extent, the presence of literals (for example specific titles of movies). Additionally, LC-QuAD 2.0 includes much more difficult query structures as suggested by the large SPARQL vocabulary size reported in Table \ref{tab:voc}. Another challenge is that some templates of LC-QuAD 2.0  share a question template.  This means that for the same question template, we can have different query templates. Since our copy mechanism is based on masking the KB vocabulary, two sentences that share the same question template will be considered exactly the same sentence by the encoder-decoder block and only the copy block will be able to "see" the words that differentiate them. 
Therefore, questions that share a question template are generated by the models following the most frequent query template associated with this question template. 
For instance, the question structure "what is the \texttt{<mask>} for \texttt{<mask>} of \texttt{<mask>}" can be generated by two global templates. The first template generates entries such as: Question: "what is the country for head of state of mahmoud abbas" / Query: \texttt{select distinct ?sbj where \{ ?sbj wdt:P35 wd:Q127998 . ?sbj wdt:P31 wd:Q6256 \}} and appears 1431 times in the train set. The second template generates entries such as: Question: "what is the medication for significant drug interaction of erythromycin" / Query: \texttt{select distinct ?obj where \{ wd:Q213511 wdt:P769 ?obj . ?obj wdt:P31 wd:Q12140 \}} and appears 1344 times in the train set. We then observe that when the model faces a question with similar structure in the test set, it will always generate a query with the first most frequent template at train time. 
This fact implies that all expected queries that do not match this query template will not be properly generated. 

We can also note that all our copy models with "tag-within" questions have very similar performances, around 73\% of accuracy.
This cap in performance does not occur for "tag-end" questions. In the "tag-end" setting, the complete NL question is passed to the models. Even though the KB elements are masked, the semantics of the question can be identified by its natural language formulation. We can see, for instance, that BART manages to outperform the cap of 70\% of accuracy with the copy mechanism on "tag-end" questions.

\paragraph{Other models.}
In this study, we used NMT techniques based on encoders-decoders that are fully available to train and test. Moreover, we only use BART and T5 for PLMs and ConvSeq2Seq and Transformer for NPLMs since we compare our results to current state-of-the-art approaches. We also tested two of the best LLMs for code generation, but there are others, such as Codex \cite{codex} and InstructGPT \cite{instructGPT}, that could be tested in the future. 


\begin{table}
\caption{ Error Distribution in \% for T5 LCQ 1 "tag-within"}
\centering
\begin{threeparttable}
\begin{tabular}{c|ccccccccc|}  
\cline{2-9}  
& \multicolumn{8}{c|}{Predictions} \\ \cline{2-9}  
\multicolumn{1}{c|}{} & \multicolumn{1}{c|}{} & \multicolumn{1}{c|}{URIs} & \multicolumn{1}{c|}{"Fake" URIs} & \multicolumn{1}{c|}{SVocab} & \multicolumn{1}{c|}{Lit} & \multicolumn{1}{c|}{Fct} & \multicolumn{1}{c|}{Var} & \multicolumn{1}{c|}{Unk} \\ \hline 
\multicolumn{1}{|c|}{\multirow{6}{*}{\rotatebox[origin=c]{90}{References}}}        & \multicolumn{1}{c|}{URIs} & \multicolumn{1}{c|}{67} & \multicolumn{1}{c|}{19} & \multicolumn{1}{c|}{11} & \multicolumn{1}{c|}{0} & \multicolumn{1}{c|}{3} & \multicolumn{1}{c|}{0} & \multicolumn{1}{c|}{0} \\ \cline{2-9}        & \multicolumn{1}{c|}{SVocab} & \multicolumn{1}{c|}{3} & \multicolumn{1}{c|}{0} & \multicolumn{1}{c|}{97} & \multicolumn{1}{c|}{0} & \multicolumn{1}{c|}{0} & \multicolumn{1}{c|}{0} & \multicolumn{1}{c|}{0} \\ \cline{2-9}        & \multicolumn{1}{c|}{Lit} & \multicolumn{1}{c|}{0} & \multicolumn{1}{c|}{0} & \multicolumn{1}{c|}{0} & \multicolumn{1}{c|}{0} & \multicolumn{1}{c|}{0} & \multicolumn{1}{c|}{0} & \multicolumn{1}{c|}{0} \\ \cline{2-9}        & \multicolumn{1}{c|}{Fct} & \multicolumn{1}{c|}{0} & \multicolumn{1}{c|}{0} & \multicolumn{1}{c|}{100} & \multicolumn{1}{c|}{0} & \multicolumn{1}{c|}{0} & \multicolumn{1}{c|}{0} & \multicolumn{1}{c|}{0} \\ \cline{2-9}        & \multicolumn{1}{c|}{Var} & \multicolumn{1}{c|}{0} & \multicolumn{1}{c|}{0} & \multicolumn{1}{c|}{0} & \multicolumn{1}{c|}{0} & \multicolumn{1}{c|}{0} & \multicolumn{1}{c|}{0} & \multicolumn{1}{c|}{0} \\ \cline{2-9}        & \multicolumn{1}{c|}{Unk} & \multicolumn{1}{c|}{0} & \multicolumn{1}{c|}{0} & \multicolumn{1}{c|}{0} & \multicolumn{1}{c|}{0} & \multicolumn{1}{c|}{0} & \multicolumn{1}{c|}{0} & \multicolumn{1}{c|}{0} \\ \cline{2-9}        & \multicolumn{8}{c|}{\textcolor{red}{URIs:33\% -- SVocab:67\% -- Lit:0\% -- Fct:0\% -- Var:0\% -- Unk:0\%}} \\ \hline       & \multicolumn{8}{c|}{\textcolor{blue}{BLEU: 99\% -- ANSWER ACC: 97\% -- ANSWER F1: 97\%}} \\ 
\hline
\end{tabular}
\begin{tablenotes}
   \item[*]  The red line shows the general distribution of each type of token, and the blue bottom highlights the model's performance in the specific setting. We have added these values again to facilitate understanding when looking at the errors by type because it is essential to look at the general performance before looking at the error distribution to properly evaluate the rates in relative values.  
  \end{tablenotes}
  \end{threeparttable}
\label{tab:errAnal1}
\end{table}

\begin{table}
\footnotesize
\centering
\caption{Error Distribution in \% for T5 LCQ 2 "tag-within"}
\begin{tabular}{c|ccccccccc|}  
\cline{2-9}  
& \multicolumn{8}{c|}{Predictions} \\ \cline{2-9}  
\multicolumn{1}{c|}{} & \multicolumn{1}{c|}{} & \multicolumn{1}{c|}{URIs} & \multicolumn{1}{c|}{"Fake" URIs} & \multicolumn{1}{c|}{SVocab} & \multicolumn{1}{c|}{Lit} & \multicolumn{1}{c|}{Fct} & \multicolumn{1}{c|}{Var} & \multicolumn{1}{c|}{Unk} \\ \hline 
\multicolumn{1}{|c|}{\multirow{6}{*}{\rotatebox[origin=c]{90}{References}}}        & \multicolumn{1}{c|}{URIs} & \multicolumn{1}{c|}{31} & \multicolumn{1}{c|}{0} & \multicolumn{1}{c|}{0} & \multicolumn{1}{c|}{0} & \multicolumn{1}{c|}{0} & \multicolumn{1}{c|}{69} & \multicolumn{1}{c|}{0} \\ \cline{2-9}        & \multicolumn{1}{c|}{SVocab} & \multicolumn{1}{c|}{2} & \multicolumn{1}{c|}{0} & \multicolumn{1}{c|}{75} & \multicolumn{1}{c|}{14} & \multicolumn{1}{c|}{2} & \multicolumn{1}{c|}{6} & \multicolumn{1}{c|}{1} \\ \cline{2-9}        & \multicolumn{1}{c|}{Lit} & \multicolumn{1}{c|}{0} & \multicolumn{1}{c|}{1} & \multicolumn{1}{c|}{14} & \multicolumn{1}{c|}{79} & \multicolumn{1}{c|}{4} & \multicolumn{1}{c|}{1} & \multicolumn{1}{c|}{1} \\ \cline{2-9}        & \multicolumn{1}{c|}{Fct} & \multicolumn{1}{c|}{3} & \multicolumn{1}{c|}{0} & \multicolumn{1}{c|}{33} & \multicolumn{1}{c|}{58} & \multicolumn{1}{c|}{6} & \multicolumn{1}{c|}{0} & \multicolumn{1}{c|}{0} \\ \cline{2-9}        & \multicolumn{1}{c|}{Var} & \multicolumn{1}{c|}{61} & \multicolumn{1}{c|}{0} & \multicolumn{1}{c|}{1} & \multicolumn{1}{c|}{0} & \multicolumn{1}{c|}{0} & \multicolumn{1}{c|}{38} & \multicolumn{1}{c|}{0} \\ \cline{2-9}        & \multicolumn{1}{c|}{Unk} & \multicolumn{1}{c|}{0} & \multicolumn{1}{c|}{0} & \multicolumn{1}{c|}{43} & \multicolumn{1}{c|}{57} & \multicolumn{1}{c|}{0} & \multicolumn{1}{c|}{0} & \multicolumn{1}{c|}{0} \\ \cline{2-9}        & \multicolumn{8}{c|}{\textcolor{red}{URIs:31\% -- SVocab:12\% -- Lit:18\% -- Fct:1\% -- Var:38\% -- Unk:0\%}} \\ \hline       & \multicolumn{8}{c|}{\textcolor{blue}{BLEU: 95\% -- ANSWER ACC: 94\% -- ANSWER F1: 94\%}} \\ 
\hline
\end{tabular}
\label{tab:errAnal2}
\end{table}

\begin{table}
\caption{ Error Distribution in \% for ConvSeq2Seq Copy LCQ 1 "tag-within"}
\centering
\begin{tabular}{c|ccccccccc|}  
\cline{2-9}  
& \multicolumn{8}{c|}{Predictions} \\ \cline{2-9}  
\multicolumn{1}{c|}{} & \multicolumn{1}{c|}{} & \multicolumn{1}{c|}{URIs} & \multicolumn{1}{c|}{"Fake" URIs} & \multicolumn{1}{c|}{SVocab} & \multicolumn{1}{c|}{Lit} & \multicolumn{1}{c|}{Fct} & \multicolumn{1}{c|}{Var} & \multicolumn{1}{c|}{Unk} \\ \hline 
\multicolumn{1}{|c|}{\multirow{6}{*}{\rotatebox[origin=c]{90}{References}}}        & \multicolumn{1}{c|}{URIs} & \multicolumn{1}{c|}{59} & \multicolumn{1}{c|}{0} & \multicolumn{1}{c|}{13} & \multicolumn{1}{c|}{5} & \multicolumn{1}{c|}{0} & \multicolumn{1}{c|}{0} & \multicolumn{1}{c|}{23} \\ \cline{2-9}        & \multicolumn{1}{c|}{SVocab} & \multicolumn{1}{c|}{0} & \multicolumn{1}{c|}{0} & \multicolumn{1}{c|}{98} & \multicolumn{1}{c|}{1} & \multicolumn{1}{c|}{0} & \multicolumn{1}{c|}{0} & \multicolumn{1}{c|}{1} \\ \cline{2-9}        & \multicolumn{1}{c|}{Lit} & \multicolumn{1}{c|}{0} & \multicolumn{1}{c|}{0} & \multicolumn{1}{c|}{0} & \multicolumn{1}{c|}{0} & \multicolumn{1}{c|}{0} & \multicolumn{1}{c|}{0} & \multicolumn{1}{c|}{0} \\ \cline{2-9}        & \multicolumn{1}{c|}{Fct} & \multicolumn{1}{c|}{0} & \multicolumn{1}{c|}{0} & \multicolumn{1}{c|}{100} & \multicolumn{1}{c|}{0} & \multicolumn{1}{c|}{0} & \multicolumn{1}{c|}{0} & \multicolumn{1}{c|}{0} \\ \cline{2-9}        & \multicolumn{1}{c|}{Var} & \multicolumn{1}{c|}{0} & \multicolumn{1}{c|}{0} & \multicolumn{1}{c|}{0} & \multicolumn{1}{c|}{0} & \multicolumn{1}{c|}{0} & \multicolumn{1}{c|}{0} & \multicolumn{1}{c|}{0} \\ \cline{2-9}        & \multicolumn{1}{c|}{Unk} & \multicolumn{1}{c|}{0} & \multicolumn{1}{c|}{0} & \multicolumn{1}{c|}{0} & \multicolumn{1}{c|}{0} & \multicolumn{1}{c|}{0} & \multicolumn{1}{c|}{0} & \multicolumn{1}{c|}{0} \\ \cline{2-9}        & \multicolumn{8}{c|}{\textcolor{red}{URIs:14\% -- SVocab:76\% -- Lit:0\% -- Fct:1\% -- Var:0\% -- Unk:9\%}} \\ \hline       & \multicolumn{8}{c|}{\textcolor{blue}{BLEU: 99\% -- ANSWER ACC: 96\% -- ANSWER F1: 96\%}} \\ 
\hline
\end{tabular}
\label{tab:errAnal3}
\end{table}

\begin{table}
\caption{ Error Distribution in \% for ConvSeq2Seq Copy LCQ 2 "tag-within"}
\centering
\begin{tabular}{c|ccccccccc|}  
\cline{2-9}  
& \multicolumn{8}{c|}{Predictions} \\ \cline{2-9}  
\multicolumn{1}{c|}{} & \multicolumn{1}{c|}{} & \multicolumn{1}{c|}{URIs} & \multicolumn{1}{c|}{"Fake" URIs} & \multicolumn{1}{c|}{SVocab} & \multicolumn{1}{c|}{Lit} & \multicolumn{1}{c|}{Fct} & \multicolumn{1}{c|}{Var} & \multicolumn{1}{c|}{Unk} \\ \hline 
\multicolumn{1}{|c|}{\multirow{6}{*}{\rotatebox[origin=c]{90}{References}}}        & \multicolumn{1}{c|}{URIs} & \multicolumn{1}{c|}{64} & \multicolumn{1}{c|}{0} & \multicolumn{1}{c|}{8} & \multicolumn{1}{c|}{1} & \multicolumn{1}{c|}{0} & \multicolumn{1}{c|}{27} & \multicolumn{1}{c|}{0} \\ \cline{2-9}        & \multicolumn{1}{c|}{SVocab} & \multicolumn{1}{c|}{0} & \multicolumn{1}{c|}{0} & \multicolumn{1}{c|}{55} & \multicolumn{1}{c|}{29} & \multicolumn{1}{c|}{2} & \multicolumn{1}{c|}{14} & \multicolumn{1}{c|}{0} \\ \cline{2-9}        & \multicolumn{1}{c|}{Lit} & \multicolumn{1}{c|}{0} & \multicolumn{1}{c|}{0} & \multicolumn{1}{c|}{100} & \multicolumn{1}{c|}{0} & \multicolumn{1}{c|}{0} & \multicolumn{1}{c|}{0} & \multicolumn{1}{c|}{0} \\ \cline{2-9}        & \multicolumn{1}{c|}{Fct} & \multicolumn{1}{c|}{100} & \multicolumn{1}{c|}{0} & \multicolumn{1}{c|}{0} & \multicolumn{1}{c|}{0} & \multicolumn{1}{c|}{0} & \multicolumn{1}{c|}{0} & \multicolumn{1}{c|}{0} \\ \cline{2-9}        & \multicolumn{1}{c|}{Var} & \multicolumn{1}{c|}{41} & \multicolumn{1}{c|}{0} & \multicolumn{1}{c|}{0} & \multicolumn{1}{c|}{0} & \multicolumn{1}{c|}{0} & \multicolumn{1}{c|}{59} & \multicolumn{1}{c|}{0} \\ \cline{2-9}        & \multicolumn{1}{c|}{Unk} & \multicolumn{1}{c|}{0} & \multicolumn{1}{c|}{0} & \multicolumn{1}{c|}{0} & \multicolumn{1}{c|}{0} & \multicolumn{1}{c|}{0} & \multicolumn{1}{c|}{0} & \multicolumn{1}{c|}{0} \\ \cline{2-9}        & \multicolumn{8}{c|}{\textcolor{red}{URIs:41\% -- SVocab:28\% -- Lit:1\% -- Fct:2\% -- Var:27\% -- Unk:1\%}} \\ \hline       & \multicolumn{8}{c|}{\textcolor{blue}{BLEU: 89\% -- ANSWER ACC: 69\% -- ANSWER F1: 69\%}} \\ 
\hline
\end{tabular}
\label{tab:errAnal4}
\end{table}

\section{Conclusion and future work}
We presented a set of experiments to compare and expand upon the state-of-the-art approaches for NMT-based SPARQL query generation and examined the impact of a copy block. We compared non-pre-trained and pre-trained models. In the case of pre-trained models (BART, T5), we are the first to evaluate adding a copy layer for this task. Given the lack of homogeneous evaluation metrics in the state of the art, we also compare three datasets using the BLEU score, the accuracy, and the F1-score computed on non-empty answers returned by the generated queries. We also show the impact of question annotation on non-pre-trained and pre-trained models. 

Our results demonstrate that the copy mechanism improves the performances of non-pre-trained models by a significant margin, including in the "tag-within" setting. We also show that the copy mechanism can improve pre-trained models' performance in some cases (BART) and lower them in others (T5). We also make a detailed analysis of the errors made by all models. In copy models, errors in generating URIs are due to a wrong choice among hidden tokens (URIs) in the input. Finally, we note that even the best PLM and NPLM are not flexible to question reformulation and, thus, do not have adequate generalization capabilities.

In our future work, we plan to investigate improvements in the copy mechanism that would keep the advantage of hiding the KB vocabulary from the encoder-decoder without making the model fully blind to its semantics. In particular, we plan to study the impact of integrating knowledge graph embedding in our models.  

\section*{Acknowledgments}
We are grateful to the NSERC Discovery Grant Program, which has funded this research. The authors would also like to express their gratitude to Compute Canada (Calcul Québec) for providing computational resources.

{\appendix[Error Analysis for LLMs]
The followings tables show the detailed errors analysis type by type for Llama v2 7B and Code Llama v2 7B models and settings. All experiences are conducted over LC-QuAD 2.0.

\begin{table}[!h]
\caption{Fine-Tuning Llama v2 7B with 25\% of "tag-end" questions of the train set.}
\centering
\begin{tabular}{c|ccccccccc|}  
\cline{2-9}  
& \multicolumn{8}{c|}{Predictions} \\ \cline{2-9}  
\multicolumn{1}{c|}{} & \multicolumn{1}{c|}{} & \multicolumn{1}{c|}{URIs} & \multicolumn{1}{c|}{FakeURIs} & \multicolumn{1}{c|}{SVocab} & \multicolumn{1}{c|}{Lit} & \multicolumn{1}{c|}{Fct} & \multicolumn{1}{c|}{Var} & \multicolumn{1}{c|}{Unk} \\ \hline 
\multicolumn{1}{|c|}{\multirow{6}{*}{\rotatebox[origin=c]{90}{References}}}        & \multicolumn{1}{c|}{URIs} & \multicolumn{1}{c|}{90} & \multicolumn{1}{c|}{5} & \multicolumn{1}{c|}{2} & \multicolumn{1}{c|}{0} & \multicolumn{1}{c|}{0} & \multicolumn{1}{c|}{3} & \multicolumn{1}{c|}{0} \\ \cline{2-9}        & \multicolumn{1}{c|}{SVocab} & \multicolumn{1}{c|}{0} & \multicolumn{1}{c|}{10} & \multicolumn{1}{c|}{61} & \multicolumn{1}{c|}{3} & \multicolumn{1}{c|}{2} & \multicolumn{1}{c|}{23} & \multicolumn{1}{c|}{0} \\ \cline{2-9}        & \multicolumn{1}{c|}{Lit} & \multicolumn{1}{c|}{0} & \multicolumn{1}{c|}{4} & \multicolumn{1}{c|}{62} & \multicolumn{1}{c|}{21} & \multicolumn{1}{c|}{5} & \multicolumn{1}{c|}{8} & \multicolumn{1}{c|}{0} \\ \cline{2-9}        & \multicolumn{1}{c|}{Fct} & \multicolumn{1}{c|}{0} & \multicolumn{1}{c|}{9} & \multicolumn{1}{c|}{63} & \multicolumn{1}{c|}{6} & \multicolumn{1}{c|}{3} & \multicolumn{1}{c|}{19} & \multicolumn{1}{c|}{0} \\ \cline{2-9}        & \multicolumn{1}{c|}{Var} & \multicolumn{1}{c|}{0} & \multicolumn{1}{c|}{22} & \multicolumn{1}{c|}{51} & \multicolumn{1}{c|}{0} & \multicolumn{1}{c|}{0} & \multicolumn{1}{c|}{26} & \multicolumn{1}{c|}{0} \\ \cline{2-9}        & \multicolumn{1}{c|}{Unk} & \multicolumn{1}{c|}{0} & \multicolumn{1}{c|}{0} & \multicolumn{1}{c|}{0} & \multicolumn{1}{c|}{0} & \multicolumn{1}{c|}{0} & \multicolumn{1}{c|}{0} & \multicolumn{1}{c|}{0} \\ \cline{2-9}        & \multicolumn{8}{c|}{\textcolor{red}{URIs:78\% -- SVocab:8\% -- Lit:1\% -- Fct:0\% -- Var:14\% -- Unk:0\%}} \\ \hline       & \multicolumn{8}{c|}{\textcolor{blue}{BLEU: 55\% -- ANSWER ACC: 0.2\% -- ANSWER F1: 0.3\%}} \\ 
\hline
\end{tabular}
\end{table}

\begin{table}[!h]
\caption{Fine-Tuning Llama v2 7B with 50\% of "tag-end" questions of the train set.}
\centering
\begin{tabular}{c|ccccccccc|}  
\cline{2-9}  
& \multicolumn{8}{c|}{Predictions} \\ \cline{2-9}  
\multicolumn{1}{c|}{} & \multicolumn{1}{c|}{} & \multicolumn{1}{c|}{URIs} & \multicolumn{1}{c|}{FakeURIs} & \multicolumn{1}{c|}{SVocab} & \multicolumn{1}{c|}{Lit} & \multicolumn{1}{c|}{Fct} & \multicolumn{1}{c|}{Var} & \multicolumn{1}{c|}{Unk} \\ \hline 
\multicolumn{1}{|c|}{\multirow{6}{*}{\rotatebox[origin=c]{90}{References}}}        & \multicolumn{1}{c|}{URIs} & \multicolumn{1}{c|}{90} & \multicolumn{1}{c|}{6} & \multicolumn{1}{c|}{1} & \multicolumn{1}{c|}{0} & \multicolumn{1}{c|}{0} & \multicolumn{1}{c|}{3} & \multicolumn{1}{c|}{0} \\ \cline{2-9}        & \multicolumn{1}{c|}{SVocab} & \multicolumn{1}{c|}{0} & \multicolumn{1}{c|}{13} & \multicolumn{1}{c|}{63} & \multicolumn{1}{c|}{5} & \multicolumn{1}{c|}{3} & \multicolumn{1}{c|}{16} & \multicolumn{1}{c|}{0} \\ \cline{2-9}        & \multicolumn{1}{c|}{Lit} & \multicolumn{1}{c|}{0} & \multicolumn{1}{c|}{2} & \multicolumn{1}{c|}{56} & \multicolumn{1}{c|}{25} & \multicolumn{1}{c|}{9} & \multicolumn{1}{c|}{8} & \multicolumn{1}{c|}{0} \\ \cline{2-9}        & \multicolumn{1}{c|}{Fct} & \multicolumn{1}{c|}{0} & \multicolumn{1}{c|}{8} & \multicolumn{1}{c|}{67} & \multicolumn{1}{c|}{5} & \multicolumn{1}{c|}{5} & \multicolumn{1}{c|}{14} & \multicolumn{1}{c|}{0} \\ \cline{2-9}        & \multicolumn{1}{c|}{Var} & \multicolumn{1}{c|}{0} & \multicolumn{1}{c|}{18} & \multicolumn{1}{c|}{58} & \multicolumn{1}{c|}{0} & \multicolumn{1}{c|}{0} & \multicolumn{1}{c|}{24} & \multicolumn{1}{c|}{0} \\ \cline{2-9}        & \multicolumn{1}{c|}{Unk} & \multicolumn{1}{c|}{0} & \multicolumn{1}{c|}{0} & \multicolumn{1}{c|}{0} & \multicolumn{1}{c|}{0} & \multicolumn{1}{c|}{0} & \multicolumn{1}{c|}{0} & \multicolumn{1}{c|}{0} \\ \cline{2-9}        & \multicolumn{8}{c|}{\textcolor{red}{URIs:79\% -- SVocab:7\% -- Lit:1\% -- Fct:0\% -- Var:12\% -- Unk:0\%}} \\ \hline       & \multicolumn{8}{c|}{\textcolor{blue}{BLEU: 56\% -- ANSWER ACC: 0.3\% -- ANSWER F1: 0.3\%}} \\ 
\hline
\end{tabular}
\end{table}

\begin{table}[!h]
\caption{Instruction-Fine-Tuning Llama v2 7B with 50\% of "tag-end" questions of the train set.}
\centering
\begin{tabular}{c|ccccccccc|}  
\cline{2-9}  
& \multicolumn{8}{c|}{Predictions} \\ \cline{2-9}  
\multicolumn{1}{c|}{} & \multicolumn{1}{c|}{} & \multicolumn{1}{c|}{URIs} & \multicolumn{1}{c|}{FakeURIs} & \multicolumn{1}{c|}{SVocab} & \multicolumn{1}{c|}{Lit} & \multicolumn{1}{c|}{Fct} & \multicolumn{1}{c|}{Var} & \multicolumn{1}{c|}{Unk} \\ \hline 
\multicolumn{1}{|c|}{\multirow{6}{*}{\rotatebox[origin=c]{90}{References}}}        & \multicolumn{1}{c|}{URIs} & \multicolumn{1}{c|}{90} & \multicolumn{1}{c|}{8} & \multicolumn{1}{c|}{0} & \multicolumn{1}{c|}{0} & \multicolumn{1}{c|}{0} & \multicolumn{1}{c|}{2} & \multicolumn{1}{c|}{0} \\ \cline{2-9}        & \multicolumn{1}{c|}{SVocab} & \multicolumn{1}{c|}{0} & \multicolumn{1}{c|}{1} & \multicolumn{1}{c|}{71} & \multicolumn{1}{c|}{21} & \multicolumn{1}{c|}{5} & \multicolumn{1}{c|}{2} & \multicolumn{1}{c|}{0} \\ \cline{2-9}        & \multicolumn{1}{c|}{Lit} & \multicolumn{1}{c|}{0} & \multicolumn{1}{c|}{2} & \multicolumn{1}{c|}{80} & \multicolumn{1}{c|}{18} & \multicolumn{1}{c|}{0} & \multicolumn{1}{c|}{1} & \multicolumn{1}{c|}{0} \\ \cline{2-9}        & \multicolumn{1}{c|}{Fct} & \multicolumn{1}{c|}{0} & \multicolumn{1}{c|}{0} & \multicolumn{1}{c|}{7} & \multicolumn{1}{c|}{16} & \multicolumn{1}{c|}{0} & \multicolumn{1}{c|}{76} & \multicolumn{1}{c|}{0} \\ \cline{2-9}        & \multicolumn{1}{c|}{Var} & \multicolumn{1}{c|}{0} & \multicolumn{1}{c|}{12} & \multicolumn{1}{c|}{71} & \multicolumn{1}{c|}{0} & \multicolumn{1}{c|}{0} & \multicolumn{1}{c|}{17} & \multicolumn{1}{c|}{0} \\ \cline{2-9}        & \multicolumn{1}{c|}{Unk} & \multicolumn{1}{c|}{0} & \multicolumn{1}{c|}{0} & \multicolumn{1}{c|}{0} & \multicolumn{1}{c|}{0} & \multicolumn{1}{c|}{0} & \multicolumn{1}{c|}{0} & \multicolumn{1}{c|}{0} \\ \cline{2-9}        & \multicolumn{8}{c|}{\textcolor{red}{URIs:60\% -- SVocab:25\% -- Lit:5\% -- Fct:2\% -- Var:8\% -- Unk:0\%}} \\ \hline       & \multicolumn{8}{c|}{\textcolor{blue}{BLEU: 56\% -- ANSWER ACC: 13\% -- ANSWER F1: 13\%}} \\ 
\hline
\end{tabular}
\end{table}

\begin{table}[!h]
\caption{Instruction-Fine-Tuning Llama v2 7B with 100\% of "tag-end" questions of the train set.}
\centering
\begin{tabular}{c|ccccccccc|}  
\cline{2-9}  
& \multicolumn{8}{c|}{Predictions} \\ \cline{2-9}  
\multicolumn{1}{c|}{} & \multicolumn{1}{c|}{} & \multicolumn{1}{c|}{URIs} & \multicolumn{1}{c|}{FakeURIs} & \multicolumn{1}{c|}{SVocab} & \multicolumn{1}{c|}{Lit} & \multicolumn{1}{c|}{Fct} & \multicolumn{1}{c|}{Var} & \multicolumn{1}{c|}{Unk} \\ \hline 
\multicolumn{1}{|c|}{\multirow{6}{*}{\rotatebox[origin=c]{90}{References}}}        & \multicolumn{1}{c|}{URIs} & \multicolumn{1}{c|}{90} & \multicolumn{1}{c|}{8} & \multicolumn{1}{c|}{0} & \multicolumn{1}{c|}{0} & \multicolumn{1}{c|}{0} & \multicolumn{1}{c|}{2} & \multicolumn{1}{c|}{0} \\ \cline{2-9}        & \multicolumn{1}{c|}{SVocab} & \multicolumn{1}{c|}{0} & \multicolumn{1}{c|}{2} & \multicolumn{1}{c|}{70} & \multicolumn{1}{c|}{21} & \multicolumn{1}{c|}{5} & \multicolumn{1}{c|}{2} & \multicolumn{1}{c|}{0} \\ \cline{2-9}        & \multicolumn{1}{c|}{Lit} & \multicolumn{1}{c|}{0} & \multicolumn{1}{c|}{2} & \multicolumn{1}{c|}{80} & \multicolumn{1}{c|}{18} & \multicolumn{1}{c|}{0} & \multicolumn{1}{c|}{1} & \multicolumn{1}{c|}{0} \\ \cline{2-9}        & \multicolumn{1}{c|}{Fct} & \multicolumn{1}{c|}{0} & \multicolumn{1}{c|}{0} & \multicolumn{1}{c|}{7} & \multicolumn{1}{c|}{17} & \multicolumn{1}{c|}{0} & \multicolumn{1}{c|}{76} & \multicolumn{1}{c|}{0} \\ \cline{2-9}        & \multicolumn{1}{c|}{Var} & \multicolumn{1}{c|}{0} & \multicolumn{1}{c|}{11} & \multicolumn{1}{c|}{72} & \multicolumn{1}{c|}{0} & \multicolumn{1}{c|}{0} & \multicolumn{1}{c|}{17} & \multicolumn{1}{c|}{0} \\ \cline{2-9}        & \multicolumn{1}{c|}{Unk} & \multicolumn{1}{c|}{0} & \multicolumn{1}{c|}{0} & \multicolumn{1}{c|}{0} & \multicolumn{1}{c|}{0} & \multicolumn{1}{c|}{0} & \multicolumn{1}{c|}{0} & \multicolumn{1}{c|}{0} \\ \cline{2-9}        & \multicolumn{8}{c|}{\textcolor{red}{URIs:61\% -- SVocab:26\% -- Lit:5\% -- Fct:2\% -- Var:6\% -- Unk:0\%}} \\ \hline       & \multicolumn{8}{c|}{\textcolor{blue}{BLEU: 56\% -- ANSWER ACC: 15\% -- ANSWER F1: 13\%}} \\ 
\hline
\end{tabular}
\end{table}

\begin{table}[!h]
\caption{Fine-Tuning Code Llamav2 7B with 25\% of tag-end questions of the train set.}
\centering
\begin{tabular}{c|ccccccccc|}  
\cline{2-9}  
& \multicolumn{8}{c|}{Predictions} \\ \cline{2-9}  
\multicolumn{1}{c|}{} & \multicolumn{1}{c|}{} & \multicolumn{1}{c|}{URIs} & \multicolumn{1}{c|}{FakeURIs} & \multicolumn{1}{c|}{SVocab} & \multicolumn{1}{c|}{Lit} & \multicolumn{1}{c|}{Fct} & \multicolumn{1}{c|}{Var} & \multicolumn{1}{c|}{Unk} \\ \hline 
\multicolumn{1}{|c|}{\multirow{6}{*}{\rotatebox[origin=c]{90}{References}}}        & \multicolumn{1}{c|}{URIs} & \multicolumn{1}{c|}{96} & \multicolumn{1}{c|}{4} & \multicolumn{1}{c|}{0} & \multicolumn{1}{c|}{0} & \multicolumn{1}{c|}{0} & \multicolumn{1}{c|}{0} & \multicolumn{1}{c|}{0} \\ \cline{2-9}        & \multicolumn{1}{c|}{SVocab} & \multicolumn{1}{c|}{0} & \multicolumn{1}{c|}{1} & \multicolumn{1}{c|}{71} & \multicolumn{1}{c|}{21} & \multicolumn{1}{c|}{5} & \multicolumn{1}{c|}{2} & \multicolumn{1}{c|}{0} \\ \cline{2-9}        & \multicolumn{1}{c|}{Lit} & \multicolumn{1}{c|}{0} & \multicolumn{1}{c|}{2} & \multicolumn{1}{c|}{80} & \multicolumn{1}{c|}{18} & \multicolumn{1}{c|}{0} & \multicolumn{1}{c|}{1} & \multicolumn{1}{c|}{0} \\ \cline{2-9}        & \multicolumn{1}{c|}{Fct} & \multicolumn{1}{c|}{0} & \multicolumn{1}{c|}{0} & \multicolumn{1}{c|}{7} & \multicolumn{1}{c|}{16} & \multicolumn{1}{c|}{0} & \multicolumn{1}{c|}{76} & \multicolumn{1}{c|}{0} \\ \cline{2-9}        & \multicolumn{1}{c|}{Var} & \multicolumn{1}{c|}{0} & \multicolumn{1}{c|}{0} & \multicolumn{1}{c|}{98} & \multicolumn{1}{c|}{0} & \multicolumn{1}{c|}{0} & \multicolumn{1}{c|}{1} & \multicolumn{1}{c|}{0} \\ \cline{2-9}        & \multicolumn{1}{c|}{Unk} & \multicolumn{1}{c|}{0} & \multicolumn{1}{c|}{0} & \multicolumn{1}{c|}{0} & \multicolumn{1}{c|}{0} & \multicolumn{1}{c|}{0} & \multicolumn{1}{c|}{0} & \multicolumn{1}{c|}{0} \\ \cline{2-9}        & \multicolumn{8}{c|}{\textcolor{red}{URIs:61\% -- SVocab:26\% -- Lit:5\% -- Fct:2\% -- Var:6\% -- Unk:0\%}} \\ \hline       & \multicolumn{8}{c|}{\textcolor{blue}{BLEU: 54\% -- ANSWER ACC: 0.2\% -- ANSWER F1: 0.1\%}} \\ 
\hline
\end{tabular}
\end{table}

\begin{table}[!h]
\caption{Fine-Tuning Code Llamav2 7B with 50\% of tag-end questions of the train set.}
\centering
\begin{tabular}{c|ccccccccc|}  
\cline{2-9}  
& \multicolumn{8}{c|}{Predictions} \\ \cline{2-9}  
\multicolumn{1}{c|}{} & \multicolumn{1}{c|}{} & \multicolumn{1}{c|}{URIs} & \multicolumn{1}{c|}{FakeURIs} & \multicolumn{1}{c|}{SVocab} & \multicolumn{1}{c|}{Lit} & \multicolumn{1}{c|}{Fct} & \multicolumn{1}{c|}{Var} & \multicolumn{1}{c|}{Unk} \\ \hline 
\multicolumn{1}{|c|}{\multirow{6}{*}{\rotatebox[origin=c]{90}{References}}}        & \multicolumn{1}{c|}{URIs} & \multicolumn{1}{c|}{93} & \multicolumn{1}{c|}{5} & \multicolumn{1}{c|}{0} & \multicolumn{1}{c|}{0} & \multicolumn{1}{c|}{0} & \multicolumn{1}{c|}{2} & \multicolumn{1}{c|}{0} \\ \cline{2-9}        & \multicolumn{1}{c|}{SVocab} & \multicolumn{1}{c|}{0} & \multicolumn{1}{c|}{1} & \multicolumn{1}{c|}{71} & \multicolumn{1}{c|}{21} & \multicolumn{1}{c|}{5} & \multicolumn{1}{c|}{2} & \multicolumn{1}{c|}{0} \\ \cline{2-9}        & \multicolumn{1}{c|}{Lit} & \multicolumn{1}{c|}{0} & \multicolumn{1}{c|}{2} & \multicolumn{1}{c|}{80} & \multicolumn{1}{c|}{18} & \multicolumn{1}{c|}{0} & \multicolumn{1}{c|}{1} & \multicolumn{1}{c|}{0} \\ \cline{2-9}        & \multicolumn{1}{c|}{Fct} & \multicolumn{1}{c|}{0} & \multicolumn{1}{c|}{0} & \multicolumn{1}{c|}{7} & \multicolumn{1}{c|}{16} & \multicolumn{1}{c|}{0} & \multicolumn{1}{c|}{76} & \multicolumn{1}{c|}{0} \\ \cline{2-9}        & \multicolumn{1}{c|}{Var} & \multicolumn{1}{c|}{0} & \multicolumn{1}{c|}{14} & \multicolumn{1}{c|}{66} & \multicolumn{1}{c|}{0} & \multicolumn{1}{c|}{0} & \multicolumn{1}{c|}{20} & \multicolumn{1}{c|}{0} \\ \cline{2-9}        & \multicolumn{1}{c|}{Unk} & \multicolumn{1}{c|}{0} & \multicolumn{1}{c|}{0} & \multicolumn{1}{c|}{0} & \multicolumn{1}{c|}{0} & \multicolumn{1}{c|}{0} & \multicolumn{1}{c|}{0} & \multicolumn{1}{c|}{0} \\ \cline{2-9}        & \multicolumn{8}{c|}{\textcolor{red}{URIs:60\% -- SVocab:25\% -- Lit:5\% -- Fct:2\% -- Var:8\% -- Unk:0\%}} \\ \hline       & \multicolumn{8}{c|}{\textcolor{blue}{BLEU: 56\% -- ANSWER ACC: 0.3\% -- ANSWER F1: 0.3\%}} \\ 
\hline
\end{tabular}
\end{table}

\begin{table}[!h]
\caption{Instruction-Fine-Tuning* Code Llamav2 7B with 50\% of tag-end* questions of the train set.}
\centering
\begin{threeparttable}
\begin{tabular}{c|ccccccccc|}  
\cline{2-9}  
& \multicolumn{8}{c|}{Predictions} \\ \cline{2-9}  
\multicolumn{1}{c|}{} & \multicolumn{1}{c|}{} & \multicolumn{1}{c|}{URIs} & \multicolumn{1}{c|}{FakeURIs} & \multicolumn{1}{c|}{SVocab} & \multicolumn{1}{c|}{Lit} & \multicolumn{1}{c|}{Fct} & \multicolumn{1}{c|}{Var} & \multicolumn{1}{c|}{Unk} \\ \hline 
\multicolumn{1}{|c|}{\multirow{6}{*}{\rotatebox[origin=c]{90}{References}}}        & \multicolumn{1}{c|}{URIs} & \multicolumn{1}{c|}{94} & \multicolumn{1}{c|}{6} & \multicolumn{1}{c|}{0} & \multicolumn{1}{c|}{0} & \multicolumn{1}{c|}{0} & \multicolumn{1}{c|}{0} & \multicolumn{1}{c|}{0} \\ \cline{2-9}        & \multicolumn{1}{c|}{SVocab} & \multicolumn{1}{c|}{0} & \multicolumn{1}{c|}{1} & \multicolumn{1}{c|}{71} & \multicolumn{1}{c|}{21} & \multicolumn{1}{c|}{5} & \multicolumn{1}{c|}{2} & \multicolumn{1}{c|}{0} \\ \cline{2-9}        & \multicolumn{1}{c|}{Lit} & \multicolumn{1}{c|}{0} & \multicolumn{1}{c|}{2} & \multicolumn{1}{c|}{80} & \multicolumn{1}{c|}{18} & \multicolumn{1}{c|}{0} & \multicolumn{1}{c|}{1} & \multicolumn{1}{c|}{0} \\ \cline{2-9}        & \multicolumn{1}{c|}{Fct} & \multicolumn{1}{c|}{0} & \multicolumn{1}{c|}{0} & \multicolumn{1}{c|}{6} & \multicolumn{1}{c|}{17} & \multicolumn{1}{c|}{0} & \multicolumn{1}{c|}{76} & \multicolumn{1}{c|}{0} \\ \cline{2-9}        & \multicolumn{1}{c|}{Var} & \multicolumn{1}{c|}{0} & \multicolumn{1}{c|}{0} & \multicolumn{1}{c|}{99} & \multicolumn{1}{c|}{0} & \multicolumn{1}{c|}{0} & \multicolumn{1}{c|}{1} & \multicolumn{1}{c|}{0} \\ \cline{2-9}        & \multicolumn{1}{c|}{Unk} & \multicolumn{1}{c|}{0} & \multicolumn{1}{c|}{0} & \multicolumn{1}{c|}{0} & \multicolumn{1}{c|}{0} & \multicolumn{1}{c|}{0} & \multicolumn{1}{c|}{0} & \multicolumn{1}{c|}{0} \\ \cline{2-9}        & \multicolumn{8}{c|}{\textcolor{red}{URIs:61\% -- SVocab:26\% -- Lit:5\% -- Fct:2\% -- Var:6\% -- Unk:0\%}} \\ \hline       & \multicolumn{8}{c|}{\textcolor{blue}{BLEU: 57\% -- ANSWER ACC: 13\% -- ANSWER F1: 13\%}} \\ 
\hline
\end{tabular}
\begin{tablenotes}
     \item[*] Candidates URIs being given in the prompt to simulate tag-within setting.
  \end{tablenotes}
\end{threeparttable}
\end{table}

\begin{table}[!h]
\caption{Instruction-Fine-Tuning* Code Llamav2 7B with 100\% of tag-end* questions of the train set.}
\centering
\begin{threeparttable}
\begin{tabular}{c|ccccccccc|}  
\cline{2-9}  
& \multicolumn{8}{c|}{Predictions} \\ \cline{2-9}  
\multicolumn{1}{c|}{} & \multicolumn{1}{c|}{} & \multicolumn{1}{c|}{URIs} & \multicolumn{1}{c|}{FakeURIs} & \multicolumn{1}{c|}{SVocab} & \multicolumn{1}{c|}{Lit} & \multicolumn{1}{c|}{Fct} & \multicolumn{1}{c|}{Var} & \multicolumn{1}{c|}{Unk} \\ \hline 
\multicolumn{1}{|c|}{\multirow{6}{*}{\rotatebox[origin=c]{90}{References}}}        & \multicolumn{1}{c|}{URIs} & \multicolumn{1}{c|}{0} & \multicolumn{1}{c|}{100} & \multicolumn{1}{c|}{0} & \multicolumn{1}{c|}{0} & \multicolumn{1}{c|}{0} & \multicolumn{1}{c|}{0} & \multicolumn{1}{c|}{0} \\ \cline{2-9}        & \multicolumn{1}{c|}{SVocab} & \multicolumn{1}{c|}{0} & \multicolumn{1}{c|}{1} & \multicolumn{1}{c|}{71} & \multicolumn{1}{c|}{21} & \multicolumn{1}{c|}{5} & \multicolumn{1}{c|}{2} & \multicolumn{1}{c|}{0} \\ \cline{2-9}        & \multicolumn{1}{c|}{Lit} & \multicolumn{1}{c|}{0} & \multicolumn{1}{c|}{2} & \multicolumn{1}{c|}{80} & \multicolumn{1}{c|}{18} & \multicolumn{1}{c|}{0} & \multicolumn{1}{c|}{1} & \multicolumn{1}{c|}{0} \\ \cline{2-9}        & \multicolumn{1}{c|}{Fct} & \multicolumn{1}{c|}{0} & \multicolumn{1}{c|}{0} & \multicolumn{1}{c|}{7} & \multicolumn{1}{c|}{17} & \multicolumn{1}{c|}{0} & \multicolumn{1}{c|}{76} & \multicolumn{1}{c|}{0} \\ \cline{2-9}        & \multicolumn{1}{c|}{Var} & \multicolumn{1}{c|}{0} & \multicolumn{1}{c|}{0} & \multicolumn{1}{c|}{99} & \multicolumn{1}{c|}{0} & \multicolumn{1}{c|}{0} & \multicolumn{1}{c|}{1} & \multicolumn{1}{c|}{0} \\ \cline{2-9}        & \multicolumn{1}{c|}{Unk} & \multicolumn{1}{c|}{0} & \multicolumn{1}{c|}{0} & \multicolumn{1}{c|}{0} & \multicolumn{1}{c|}{0} & \multicolumn{1}{c|}{0} & \multicolumn{1}{c|}{0} & \multicolumn{1}{c|}{0} \\ \cline{2-9}        & \multicolumn{8}{c|}{\textcolor{red}{URIs:61\% -- SVocab:26\% -- Lit:5\% -- Fct:2\% -- Var:6\% -- Unk:0\%}} \\ \hline       & \multicolumn{8}{c|}{\textcolor{blue}{BLEU: 59\% -- ANSWER ACC: 23\% -- ANSWER F1: 23\%}} \\ 
\hline
\end{tabular}
\begin{tablenotes}
     \item[*] Candidates URIs being given in the prompt to simulate tag-within setting.
  \end{tablenotes}
\end{threeparttable}
\end{table}

\newpage
\bibliographystyle{IEEEtran}
\bibliography{biblio}

\vspace{11pt}

\vfill

\end{document}